\documentclass[Afour,sageh,times]{sagej}

\usepackage{enumitem}                               % [inline]
\usepackage{extarrows}                              % http://ctan.org/pkg
\usepackage{amsmath,amssymb,amsfonts,dsfont} % math
\usepackage{mathtools}
\usepackage{pifont}% http://ctan.org/pkg/pifont
\usepackage{footnote}
\usepackage[perpage]{footmisc}
\makesavenoteenv{tabular}
\makesavenoteenv{table}

\usepackage[linesnumbered,ruled,vlined]{algorithm2e}
\usepackage{graphicx,tabularx,subcaption}
\graphicspath{{fig/}}
\usepackage[export]{adjustbox}% http://ctan.org/pkg/adjustbox
\usepackage{makecell,booktabs}
\usepackage[titletoc]{appendix}
\usepackage[colorlinks,bookmarksopen,bookmarksnumbered,citecolor=red,urlcolor=red]{hyperref}

\def\argmax{\mathop{\arg\max}\limits}

\newcommand{\indicator}{\mathds{1}}

\newcommand{\scaleMathLine}[2][1]{\resizebox{#1\linewidth}{!}{$\displaystyle{#2}$}}
\newcommand{\norm}[1]{\left\lVert#1\right\rVert}
\newcommand{\prl}[1]{\left(#1\right)}
\newcommand{\brl}[1]{\left[#1\right]}
\newcommand{\crl}[1]{\left\{#1\right\}}

\newcommand{\defeq}{\vcentcolon=}

\def\etal/{et~al.}
\renewcommand{\vec}[1]{\boldsymbol{#1}}

\newtheorem{proposition}{Proposition}
\newtheorem{lemma}{Lemma}

\theoremstyle{definition}
\newtheorem{problem}{Problem}
\newtheorem{definition}{Definition}

\newcommand{\BibTeX}{{\rmfamily B\kern-.05em \textsc{i\kern-.025em b}\kern-.08em
T\kern-.1667em\lower.7ex\hbox{E}\kern-.125emX}}

\def\volumeyear{2020}

\begin{document}

\runninghead{Wang et al.}

\title{Inverse reinforcement learning for autonomous navigation via differentiable semantic mapping and planning}

\author{Tianyu Wang\affilnum{1}, Vikas Dhiman\affilnum{1}, Nikolay Atanasov\affilnum{1}}

\affiliation{\affilnum{1}Department of Electrical and Computer Engineering, University of California San Diego, La Jolla, CA 92093.}

\corrauth{Tianyu Wang, Department of Electrical and Computer Engineering, University of California San Diego, La Jolla, CA 92093.}

\email{tiw161@eng.ucsd.edu}

\begin{abstract}
This paper focuses on inverse reinforcement learning for autonomous 
navigation using distance and semantic category observations. 
The objective is to infer a cost function that explains demonstrated behavior 
while relying only on the expert's observations and state-control trajectory. 
We develop a map encoder, that infers semantic category probabilities from the 
observation sequence, and a cost encoder, defined as a deep neural network over 
the semantic features. 
Since the expert cost is not directly observable, the model parameters can 
only be optimized by differentiating the error between demonstrated controls and 
a control policy computed from the cost estimate. 
We propose a new model of expert behavior that enables error minimization using 
a closed-form subgradient computed only over a subset of promising states via 
a motion planning algorithm.
Our approach allows generalizing the learned behavior to new environments with new 
spatial configurations of the semantic categories. 
We analyze the different components of our model in a minigrid environment.
We also demonstrate that our approach learns to follow traffic rules 
in the autonomous driving CARLA simulator by relying on semantic observations of 
buildings, sidewalks, and road lanes.
\end{abstract}

\keywords{Inverse reinforcement learning, semantic mapping, autonomous navigation}

\maketitle

\section{Introduction}
\label{sec:introduction}

% 1. Q: Start with the broadest base, a well accepted problem or technique
% 2. Q: What are the typical/SOTA ways of solving that problem.
% 3. Q: What are the limitations of SOTA ways solving problem
% 4. Q: What do you propose
% 5. Q: How does it solve the problem

% 1. Setup robot navigation/IRL problem
Autonomous systems operating in unstructured, partially observed, and 
changing real-world environments need an understanding of semantic meaning 
to evaluate the safety, utility, and efficiency of their performance. 
For example, while a bipedal robot may navigate along sidewalks, 
an autonomous car needs to follow the road lanes and traffic signs. 
Designing a cost function that encodes such rules by hand is infeasible for complex tasks. 
It is, however, often possible to obtain demonstrations of desirable behavior 
that indirectly capture the role of semantic context in the task execution. 
Semantic labels provide rich information about the relationship 
between object entities and their utility for task execution.
In this work, we consider an inverse reinforcement learning (IRL) problem in which 
observations containing semantic information about the environment are available.

\begin{figure}
  \centering
  \includegraphics[width=\linewidth,trim=0mm 14mm 0mm 0mm, clip]{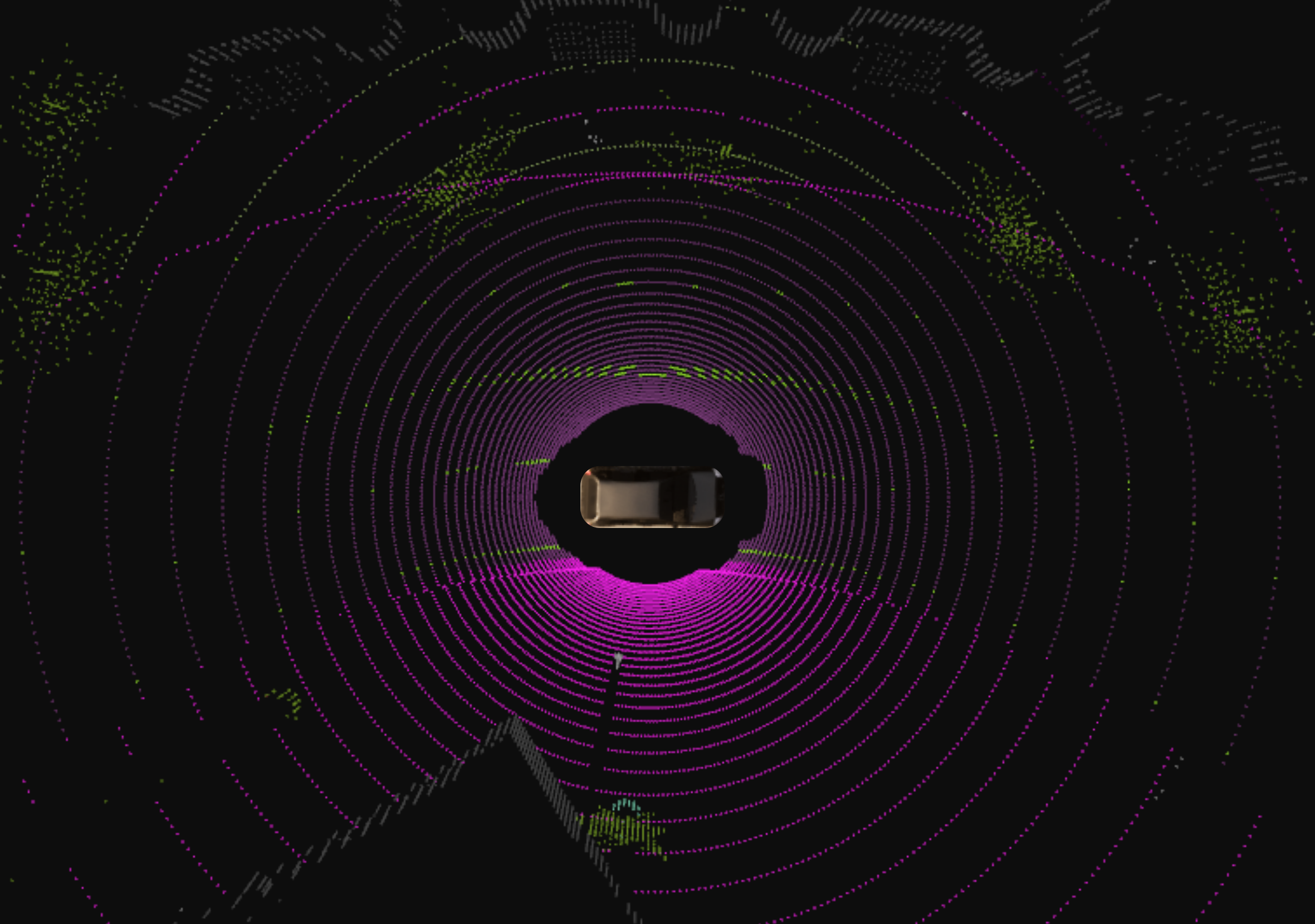}
  \caption{Autonomous vehicle in an urban street, simulated via the CARLA simulator \citep{Dosovitskiy2017CARLA}. The vehicle is equipped with a LiDAR scanner, four RGB cameras, and a segmentation algorithm, providing a semantically labeled point cloud. An expert driver demonstrates lane keeping (green) and avoidance of sidewalks (pink) and buildings (gray). This paper considers inferring the expert's cost function and generating behavior that can imitate the expert's response to semantic observations in new operational conditions.}
  \label{fig:carla_simulator}
\end{figure}

% 2. Problem description
Consider imitating a driver navigating in an unknown environment as a motivating scenario 
(see Fig.~\ref{fig:carla_simulator}).
The car is equipped with sensors that can reveal information about the semantic 
categories of surrounding objects and areas.
An expert driver can reason about a course of action based on this contextual information.
For example, staying on the road relates to making progress, 
while hitting the sidewalk or a tree should be avoided.
One key challenge in IRL is to infer a cost function when such expert 
reasoning is not explicit.
If reasoning about semantic entities can be learned from the expert demonstrations,
the cost model may generalize to new environment configurations.
To this end, we propose an IRL algorithm that learns a cost function from 
semantic features of the environment. 
Simultaneously recognizing the environment semantics and encoding costs over them is 
a very challenging task.
While other works learn a black-box neural network parametrization to map observations 
directly to costs \citep{Wulfmeier2016DeepMaxEnt, Song2019IRL}, we take advantage of 
semantic segmentation and occupancy mapping before inferring the cost function.
A metric-semantic map is constructed from causal partial semantic observations
of the environment to provide features for cost function learning.
Contrary to most IRL algorithms, which are based on the maximum entropy expert 
model \citep{Ziebart2008MaxEnt, Wulfmeier2016DeepMaxEnt}, we propose a new expert 
model allowing bounded rational deviations from optimal behavior \citep{Baker2007Goal}. 
Instead of dynamic programming over the entire state space, our formulation allows 
efficient deterministic search over a subset of promising states. 
A key advantage of our approach is that this deterministic planning process can 
be differentiated in closed-form with respect to the parameters of the learnable cost function.

%Contrary to most IRL algorithms, which are based on the maximum entropy expert model \citep{Ziebart2008MaxEnt, Wulfmeier2016DeepMaxEnt}, our proposed method does not perform dynamic programming on the entire state space 
%to bootstrap a value function for planning.
%Instead, we propose to use a noisy-rational Boltzmann distribution policy 
%to model the expert behavior~\citep{Baker2007Goal}, which allows us to perform 
%an efficient search over a subset of promising states with the learnable cost function 
%and differentiate through the planning component in closed-form. 

% 3. Our contribution
This work makes the following contributions:
\begin{enumerate}

    \item We propose a cost function representation composed of a \textit{map encoder}, 
capturing semantic class probabilities from online, first-person, distance and semantic
observations and a \textit{cost encoder}, defined as a deep neural network over 
the semantic features.
    
    \item We propose a new expert model which enables cost parameter optimization with a closed-form subgradient of the cost-to-go, computed only over a subset of promising states.
    
    \item We evaluate our model in autonomous navigation experiments in a 2D minigrid environment~\citep{gym_minigrid} with multiple semantic categories (e.g. wall, lawn, lava) as well as an autonomous driving task that respects traffic rules in the CARLA simulator~\citep{Dosovitskiy2017CARLA}.

\end{enumerate}

%The remainder of this paper is organized as follows. 
%Section~\ref{sec:related_work} examines this work in the context of existing literature 
%in inverse reinforcement learning, and semantic mapping. 
%Section~\ref{sec:problem_formulation} formally frames the problem of learning 
%a cost function from semantic observations. 
%Section~\ref{sec:cost_representation} and~\ref{sec:cost_learning} presents the main
%contributions of this paper: how to represent a cost function from streaming observations 
%with semantic information and how to optimize the cost function such that 
%it mimics the expert demonstrations. 
%Section~\ref{sec:minigrid_experiment} showcases learning navigation cost 
%in simple grid environments. 
%Section~\ref{sec:carla_experiment} demonstrates an application in the CARLA 
%autonomous driving simulation environment. 

\section{Related Work}
\label{sec:related_work}

% IL
\subsection{Imitation learning}
Imitation learning (IL) has a long history in reinforcement learning and robotics \citep{Ross2011DAgger, Atkeson1997robot, Argall2009Survey, Pastor2009Learning, Zhu2018RSS, Rajeswaran2018RSS, Pan2020IL}. %
The goal is to learn a mapping from observations to a control policy to mimic expert demonstrations. 
Behavioral cloning \citep{Ross2011DAgger} is a supervised learning approach that directly
maximizes the likelihood of the expert demonstrated behavior.
However, it typically suffers from distribution mismatch between 
training and testing and does not consider long-horizon planning. 
Another view of IL is through inverse reinforcement learning where the learner
recovers a cost function under which the expert is 
optimal \citep{Neu2012Apprenticeship, Ng2000IRL, Abbeel2004IRL}.
Recently, \citet{Ghasemipour2020divergence} and \citet{Ke2020imitation} independently developed
a unifying probabilistic perspective for common IL algorithms using various 
f-divergence metrics between the learned and expert policies as minimization objectives.  
For example, behavioral cloning minimizes the Kullback-Leibler (KL) divergence between
the learner and expert policy distribution while adversarial training methods, such as AIRL \citep{Fu2018AIRL} and GAIL \citep{Ho2016GAIL} minimize the KL divergence and Jenson Shannon divergence, respectively, between state-control distributions 
under the learned and expert policies.

% IRL
\subsection{Inverse reinforcement learning}
Learning a cost function from demonstration requires a control policy that is differentiable 
with respect to the cost parameters. 
Computing policy derivatives has been addressed by several successful IRL 
approaches \citep{Neu2012Apprenticeship, Ratliff2006MMP, Ziebart2008MaxEnt}. 
Early works assume that the cost is linear in the feature vector and aim at matching the 
feature expectations of the learned and expert policies. 
\citet{Ratliff2006MMP} compute subgradients of planning algorithms to guarantee that
the expected reward of an expert policy is better than any other policy by a margin. 
Value iteration networks (VIN) by \citet{Tamar2016VIN} show that the value iteration algorithm 
can be approximated by a series of convolution and maxpooling layers, allowing automatic 
differentiation to learn the cost function end-to-end. 
\citet{Ziebart2008MaxEnt} develop a dynamic programming algorithm to maximize the likelihood 
of observed expert data and learn a policy with maximum entropy (MaxEnt). 
Many works \citep{Levine2011Nonlinear, Wulfmeier2016DeepMaxEnt, Song2019IRL} extend MaxEnt to 
learn a nonlinear cost function using Gaussian Processes or deep neural networks. 
\citet{Finn2016GCL} use a sampling-based approximation of the MaxEnt partition function
to learn the cost function under unknown dynamics for high-dimensional 
continuous systems. 
However, the cost in most existing work is learned offline using full observation sequences 
from the expert demonstrations. 
A major contribution of our work is to develop cost representations and planning algorithms 
that rely only on causal partial observations.

% mapping and irl
\subsection{Mapping and planning}
There has been significant progress in semantic segmentation techniques, 
including deep neural networks for RGB image 
segmentation \citep{Papandreou2015WeakSemSeg,Badrinarayanan2017Segnet, Chen2018EncoderSemSeg} 
and point cloud labeling via spherical depth 
projection \citep{Wu2018SqueezeSeg, Dohan2015LidarSemSeg, Milioto2019RangeNet++,Cortinhal2020SalsaNext}. 
Maps that store semantic information can be generated from segmented 
images \citep{Sengupta2012SemMap, Lu2019MonoSemMap}. 
\citet{Gan2019Bayesian} and \citet{Sun2018ReccOctoMap} generalize binary occupancy mapping \citep{Hornung2013Octomap} to multi-class semantic mapping in 3D. In this work, we parameterize the navigation cost of an autonomous vehicle as a nonlinear 
function of such semantic map features to explain expert demonstrations.

Achieving safe and robust navigation is directly coupled with the quality of the environment 
representation and the cost function specifying desirable behaviors. 
Traditional approaches combine geometric mapping of occupancy probability \citep{Hornung2013Octomap} or distance to 
the nearest obstacle \citep{Oleynikova2017Voxblox} with hand-specified planning cost functions. 
Recent advances in deep reinforcement learning demonstrated that control inputs may be 
predicted directly from sensory observations \citep{visuomotor}. 
However, special model designs \citep{Khan2018MACN} that serve as a latent map are needed 
in navigation tasks where simple reactive policies are not feasible. 
\citet{Gupta2017CMP} decompose visual navigation into two separate stages explicitly: 
mapping the environment from first-person RGB images in local coordinates 
and planning through the constructed map with VIN \citep{Tamar2016VIN}. 
Our model constructs a global map instead and, yet, remains scalable with the size of the environment due to our sparse tensor implementation.

This paper is a revised and extended version of our previous conference publications \citep{Wang2020ICRA, Wang2020L4DC}.
In our previous work \citep{Wang2020ICRA}, we proposed differentiable mapping and planning stages to learn the expert cost function. 
The cost function is parameterized as a neural network over binary occupancy probabilities, updated from local distance observations. 
An A* motion planning algorithm computes the policy at the current state and backpropagates the gradient in closed-form to optimize the cost parameterization. We proposed an extension of the occupancy map from binary to multi-class in \citet{Wang2020L4DC}, which allows the cost function to capture semantic features from the environment. 
This paper unifies our results in a common differentiable multi-class mapping and planning architecture and presents an in-depth analysis of the various model components via experiments in a minigrid environment and the CARLA simulator. This work also introduces a new sparse tensor implementation of the multi-class occupancy mapping stage to enable its use in large environments.

\section{Problem Formulation}
\label{sec:problem_formulation}

\subsection{Environment and agent models}
\begin{figure}
\centering
\includegraphics[width=0.9\linewidth]{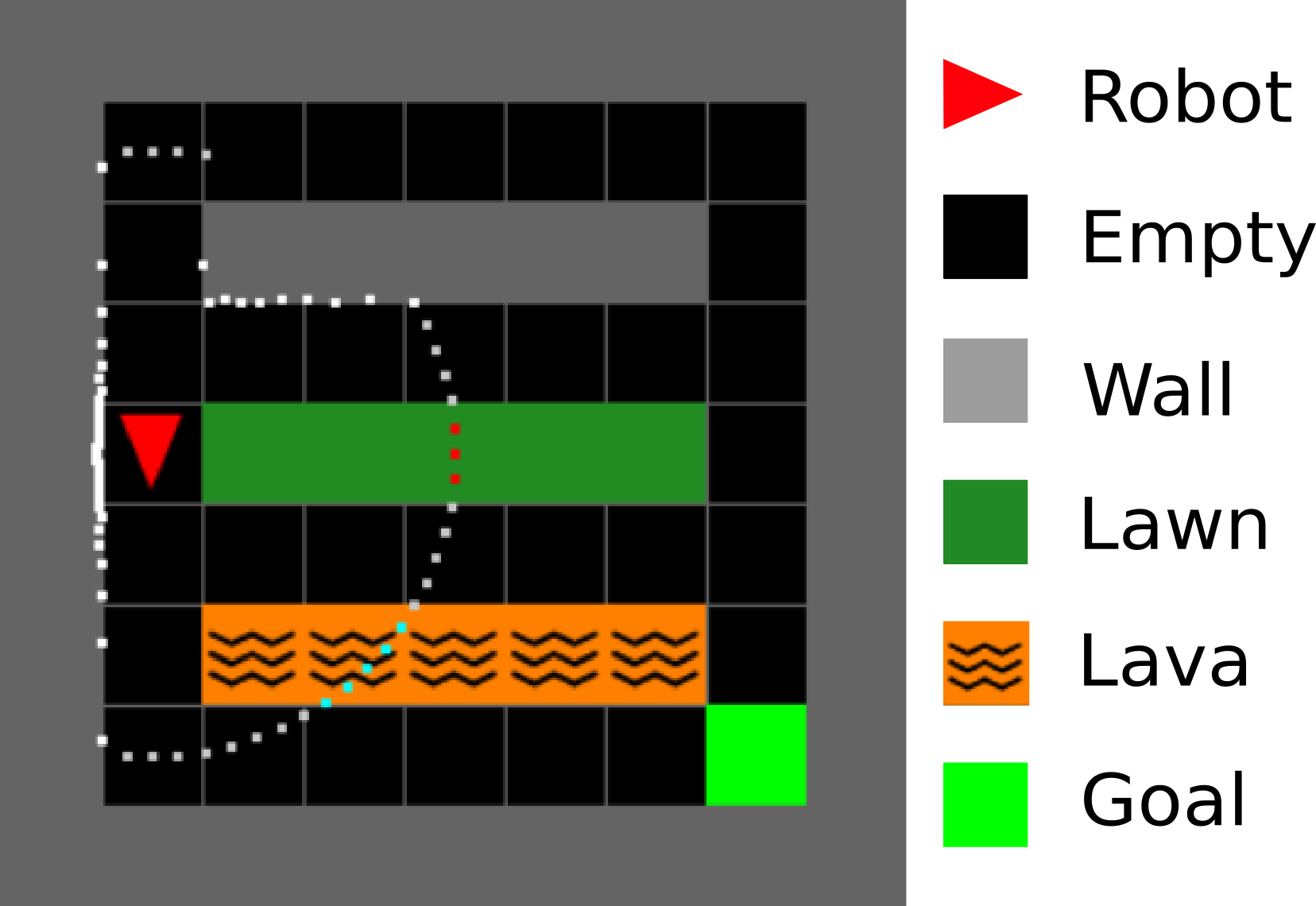}
\caption{A $9\times 9$ grid environment with cells from four semantic classes: empty, wall, lawn, lava. An autonomous agent (red triangle, facing down) starts from the top left corner and is heading towards the goal in the bottom right. The agent prefers traversing the lawn but dislikes lava. LiDAR points detect the semantic labels of the corresponding tiles (gray on empty, white on wall, purple on lawn and cyan on lava).}
\label{fig:minigrid_env}
\end{figure}

Consider an agent aiming to reach a goal in an a priori unknown environment with different terrain types. Fig.~\ref{fig:minigrid_env} shows a grid-world illustration of this setting. Let $\vec{x}_t \in \mathcal{X}$ denote the agent state (e.g., pose, twist, etc.) at discrete time $t$. Let $\vec{x}_g \in \mathcal{X}$ be the goal state. The agent state evolves according to known deterministic dynamics, $\vec{x}_{t+1} = f(\vec{x}_t, \vec{u}_t)$, with control input $\vec{u}_t \in \mathcal{U}$. The control space $\mathcal{U}$ is assumed finite. Let $\mathcal{K} = \crl{0,1,2,\dots,K}$ be a set of class labels, where $0$ denotes ``free'' space and $k \in \mathcal{K} \setminus \crl{0}$ denotes a particular semantic class such as road, sidewalk, or car. Let $m^* : \mathcal{X} \rightarrow \mathcal{K}$ be a function specifying the \textit{true} semantic occupancy of the environment by labeling states with semantic classes. We implicitly assume that $m^*$ assigns labels to agent positions rather than to other state variables. We do not introduce an output function, mapping an agent state to its position, to simplify the notation. Let $\mathcal{M}$ be the space of possible environment realizations $m^*$. Let $c^* : \mathcal{X} \times \mathcal{U} \times \mathcal{M} \rightarrow \mathbb{R}_{\ge 0}$ be a cost function specifying desirable agent behavior in a given environment, e.g., according to an expert user or an optimal design. We assume that the agent does not have access to either the true semantic map $m^*$ or the true cost function $c^*$. However, the agent is able to obtain point-cloud observations $\vec{P}_t = \crl{\prl{\vec{p}_l, \vec{y}_l}}_l \in \mathcal{P}$ at each step $t$,  where $\vec{p}_l \in \mathcal{X}$ is the measurement location and $\vec{y}_l = \brl{y_l^1, \dots, y_l^K}^\top$ is a vector of weights $y_l^k \in \mathbb{R}$, indicating the likelihood that semantic class $k \in \mathcal{K} \setminus \crl{0}$ was observed. For example, $\vec{y}_l \in \mathbb{R}^K$ can be obtained from the softmax output of a semantic segmentation algorithm \citep{Papandreou2015WeakSemSeg, Badrinarayanan2017Segnet, Chen2018EncoderSemSeg} that predicts the semantic class of the corresponding measurement location $\vec{p}_l$ in an RGBD image. The observed point cloud $\vec{P}_t$ depends on the agent state $\vec{x}_t$ and the environment realization $m^*$.

%such that 
%$\vec{y}_l = \brl{y_l^1, \dots, y_l^K}^T, y_l^k \geq 0, \sum_{k=1}^K y_l^k = 1$, 
%whose support is $\mathcal{K} \setminus \crl{0}$. In practice, $\vec{y}_l$ can be obtained from a semantic segmentation 
%algorithm~\citep{Papandreou2015WeakSemSeg, Badrinarayanan2017Segnet, Chen2018EncoderSemSeg}
%that predicts the semantic class of the corresponding measurement location $\vec{p}_l$.

% python3 src/visualize_expert.py --render --env MiniGrid-LavaLawnS9-v0

\begin{figure*}
  \centering
  \includegraphics[width=\linewidth,trim=0mm 3mm 0mm 3mm, clip]{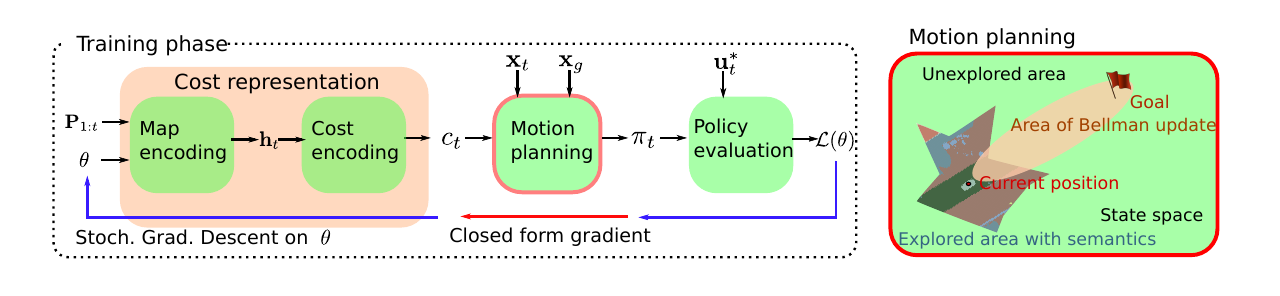}
  \caption{Architecture for cost function learning from demonstrations with semantic observations. Our main contribution is a cost representation, combining a probabilistic \emph{semantic map encoder}, with recurrent dependence on semantic observations $\vec{P}_{1:t}$, and a \emph{cost encoder}, defined over the semantic features $\vec{h}_t$. Efficient forward policy computation and closed-form subgradient backpropagation are used to optimize the cost representation parameters $\vec{\theta}$ in order to explain the expert behavior.}
  \label{fig:approach}
\end{figure*}

\subsection{Expert model}
\label{sec:expert_model}
%We assume the expert chooses a control according to a Boltzmann-rational policy
%\citep{Ramachandran2007BayesianIRL, Neu2012Apprenticeship} given the \textit{true} 
%cost $c^*$ and the \textit{true} environment $m^*$,

We assume that an expert user or algorithm demonstrates desirable agent behavior in the form of a training set $\mathcal{D} := \crl{(\vec{x}_{t,n},\vec{u}_{t,n}^*,\vec{P}_{t,n}, \vec{x}_{g,n})}_{t=1, n=1}^{T_n, N}$. The training set consists of $N$ demonstrated executions with different lengths $T_n$ for $n \in \crl{1,\ldots,N}$. Each demonstration trajectory contains the agent states $\vec{x}_{t,n}$, expert controls $\vec{u}_{t,n}^*$, and sensor observations $\vec{P}_{t,n}$ encountered during navigation to a goal state $\vec{x}_{g,n}$.

The design of an IRL algorithm depends on a model of the stochastic control policy $\pi^*(\vec{u} | \vec{x}; c^*, m^*)$ used by the expert to generate the training data $\mathcal{D}$, given the \textit{true} cost $c^*$ and environment $m^*$. The state of the art relies on the MaxEnt model \citep{Ziebart2008MaxEnt}, which assumes that the expert minimizes the weighted sum of the stage cost $c^*(\vec{x}, \vec{u} ; m^*)$ and the negative policy entropy over the agent trajectory.
% $-\mathcal{H}(\pi(\cdot | \vec{x}))$ 

We propose a new model of expert behavior to explain rational deviation from optimality. We assume that the expert is aware of the optimal value function:
\begin{align}
\label{eq:Q_star}
Q^*(\vec{x}_t, \vec{u}_t; c^*, m^*) := &\min_{T,\vec{u}_{t+1:T-1}} \sum_{k=t}^{T-1} c^*(\vec{x}_k, \vec{u}_k ; m^*) \\
&\;\;\text{s.t.}\;\; \vec{x}_{k+1} = f(\vec{x}_k, \vec{u}_k),\; \vec{x}_T = \vec{x}_g. \notag
\end{align}
but does not always choose strictly rational actions. Instead, the expert behavior is modeled as a Boltzmann policy over the optimal value function:
\begin{equation}
\label{eq:boltzmann_expert_policy}
  \pi^*(\vec{u}_t | \vec{x}_t; c^*, m^*) \!= \!
  \frac{\exp(-\frac{1}{\alpha} Q^*(\vec{x}_t, \vec{u}_t; c^*, m^*))}
  {\sum_{\vec{u} \in \mathcal{U}}\exp(-\frac{1}{\alpha} Q^*(\vec{x}_t, \vec{u}; c^*, m^*))}
\end{equation}
where $\alpha$ is a temperature parameter. 
The Boltzmann policy stipulates an exponential preference of controls that incur low long-term costs. We will show in Sec.~\ref{sec:cost_learning} that this expert model allows very efficient policy search as well as computation of the policy gradient with respect to the stage cost, which is needed for inverse cost learning. In contrast, the MaxEnt policy requires either value iteration over the full state space \citep{Ziebart2008MaxEnt} or sampling-based estimation of a partition function \citep{Finn2016GCL}. Appendix \ref{sec:comparison_boltzmann_maxent_policies} provides a comparison between our model and the MaxEnt formulation.

\subsection{Problem statement}

Given the training set $\mathcal{D}$, our goal is to:
\begin{itemize}[leftmargin=*]
    \item learn a cost function estimate $c_t: \mathcal{X} \times \mathcal{U} \times 
    \mathcal{P}^t \times \Theta \rightarrow \mathbb{R}_{\ge 0}$ 
    that depends on an observation sequence $\vec{P}_{1:t}$ from the true latent
    environment and is parameterized by $\vec{\theta} \in \Theta$,
    \item design a stochastic policy $\pi_t$ from $c_t$ such that the agent behavior 
    under $\pi_t$ matches the demonstrations in $\mathcal{D}$.
\end{itemize}
The optimal value function corresponding to a stage cost estimate $c_t$ is:
\begin{align}
\label{eq:Q}
Q_t(\vec{x}_t, \vec{u}_t; \vec{P}_{1:t}, \vec{\theta}) := & \!\!\!\!\min_{T,\vec{u}_{t+1:T-1}} \sum_{k=t}^{T-1} c_t(\vec{x}_k, \vec{u}_k ; \vec{P}_{1:t}, \vec{\theta})\!\!\\
&\;\;\text{s.t.}\;\; \vec{x}_{k+1} = f(\vec{x}_k, \vec{u}_k),\; \vec{x}_T = \vec{x}_g. \notag
\end{align}
Following the expert model proposed in Sec.~\ref{sec:expert_model}, we define a Boltzmann policy corresponding to $Q_t$:
\begin{equation}
\label{eq:boltzmann_agent_policy}
  \pi_t(\vec{u}_t | \vec{x}_t; \vec{P}_{1:t}, \vec{\theta}) \propto
  \exp(-\frac{1}{\alpha} Q_{t}(\vec{x}_t, \vec{u}_t; \vec{P}_{1:t}, \vec{\theta}))
\end{equation}
and aim to optimize the stage cost parameters $\vec{\theta}$ to match the demonstrations in $\mathcal{D}$.

\begin{problem}
\label{pb:1}
Given demonstrations $\mathcal{D}$, optimize the cost function 
parameters $\vec{\theta}$ so that log-likelihood of the demonstrated 
controls $\vec{u}_{t,n}^*$ is maximized by policy functions $\pi_{t,n}$ obtained according to \eqref{eq:boltzmann_agent_policy}:
\begin{equation}
\label{eq:loss}
\min_{\vec{\theta}} \mathcal{L}(\vec{\theta}) \defeq 
- \sum_{n=1}^N \sum_{t=1}^{T_n}\log \pi_{t, n} (\vec{u}_{t,n}^*|\vec{x}_{t,n};
\vec{P}_{1:t, n}, \vec{\theta}).
\end{equation}
\end{problem}

The problem setup is illustrated in Fig.~\ref{fig:approach}. 
An important consequence of our expert model is that the computation of the optimal 
value function corresponding to a given stage cost estimate is a standard deterministic 
shortest path (DSP) problem \citep{Bertsekas1995OC}. 
However, the challenge is to make the value function computation differentiable with 
respect to the cost parameters $\vec{\theta}$ in order to propagate the loss in 
\eqref{eq:loss} back through the DSP problem to update $\vec{\theta}$. 
Once the parameters are optimized, the associated agent behavior can be generalized 
to navigation tasks in new partially observable environments by evaluating the 
cost $c_t$ based on the observations $\vec{P}_{1:t}$ iteratively and 
re-computing the associated policy $\pi_t$.

\section{Cost Function Representation}
\label{sec:cost_representation}
% Map parameters
\newcommand{\tm}{\vec{\Psi}}
% Map parameter at m-th ray
\newcommand{\tmm}{\tm_l}
% Cost Parameters
\newcommand{\tc}{\vec{\phi}}

%\begin{figure}[t]
%  \centering
%  \def\svgwidth{0.9\linewidth}
%  \input{fig/CostNeuralNet.pdf_tex}
%  %\includegraphics[width=\linewidth]{fig/CostNeuralNet}
%  \caption{Neural network model of a cost function. A recurrent neural network, parameterized by $\tm$, takes in sequential observations $\vec{P}_{1:t}$ and outputs a latent map representation. A convolutional neural network, parameterized by $\tc$, extracts features from the map state to specify the cost $c_t$ at a given robot state-control pair $(\vec{x},\vec{u})$. The learnable parameters are $\vec{\theta} = \crl{\tm, \tc}$.}
%  \label{fig:cost_neural_net}
%\end{figure}

We propose a cost function representation with two components: a \emph{semantic occupancy map encoder} with parameters $\tm$ and a \emph{cost encoder} with parameters $\tc$. The model is differentiable by design, allowing its parameters to be optimized by the subsequent planning algorithm described in Sec.~\ref{sec:cost_learning}.

\subsection{Semantic occupancy map encoder}
\label{sec:map_encoder}

We develop a semantic occupancy map that stores the likelihood of the different semantic categories in $\mathcal{K}$ in different areas of the map. We discretize the state space $\mathcal{X}$ into $J$ cells and let $\vec{m} = \brl{m^1, \dots, m^J}^\top \in \mathcal{K}^J$ be an a priori unknown vector of true semantic labels over the cells. Given the agent states $\vec{x}_{1:t}$ and observations $\vec{P}_{1:t}$ over time, our model maintains the semantic occupancy posterior $\mathbb{P}(\vec{m} = \vec{k} | \vec{x}_{1:t}, \vec{P}_{1:t})$, where $\vec{k} = \brl{k^1, \dots, k^J}^\top \in \mathcal{K}^J$. The representation complexity may be reduced significantly if one assumes independence among the map cells $m^j$: $\mathbb{P}(\vec{m} = \vec{k} | \vec{x}_{1:t}, \vec{P}_{1:t}) =  \prod_{j=1}^J \mathbb{P}(m^j = k^j | \vec{x}_{1:t}, \vec{P}_{1:t})$.

We generalize the binary occupancy grid mapping algorithm \citep{Thrun2005PR,Hornung2013Octomap} to obtain incremental Bayesian updates for the mutli-class probability at each cell $m^j$. In detail, at time $t-1$, we maintain a vector $\vec{h}_{t-1,j}$ of class log-odds at each cell and update them given the observation $\vec{P}_{t}$ obtained from state $\vec{x}_{t}$ at time $t$.

\begin{definition}
The vector of class log-odds associated with cell $m^j$ at time $t$ is $\vec{h}_{t,j} = \brl{h_{t,j}^0, \dots, h_{t,j}^K}^\top$ with elements:
\begin{equation}
\label{eq:log_odds}
h_{t,j}^k \defeq \log \frac{\mathbb{P}(m^j = 
k | \vec{x}_{1:t}, \vec{P}_{1:t})}{\mathbb{P}(m^j = 0 | \vec{x}_{1:t}, \vec{P}_{1:t})}
\;\text{for}\; k \in \mathcal{K} \;.
\end{equation}
\end{definition}

Note that by definition, $h_{t,j}^0 = 0$. Applying Bayes rule to \eqref{eq:log_odds} leads to a recursive Bayesian update for the log-odds vector:
\begin{align}
\label{eq:bayesian_update}
&h_{t,j}^k = h_{t-1,j}^k  + \log\frac{p(\vec{P}_t | m^j = k, \vec{x}_t)}{p(\vec{P}_t | m^j = 0, \vec{x}_{t})}\\
&= h_{t-1,j}^k  + \!\!\!\sum_{\prl{\vec{p}_l, \vec{y}_l} \in \vec{P}_{t}} \!\! \prl{\log\frac{\mathbb{P}(m^j = k | \vec{x}_t, (\vec{p}_l, \vec{y}_l))}{\mathbb{P}(m^j = 0| \vec{x}_{t},(\vec{p}_l, \vec{y}_l))}  - h_{0,j}^k}, \notag
\end{align}
where we assume that the observations $\prl{\vec{p}_l, \vec{y}_l} \in \vec{P}_{t}$ at time $t$, given the cell $m^j$ and state $\vec{x}_t$, are independent among each other and of the previous observations $\vec{P}_{1:t-1}$. The semantic class posterior can be recovered from the log-odds vector $\vec{h}_{t,j}$ via a softmax function 
$\mathbb{P}(m^j = k | \vec{x}_{1:t}, \vec{P}_{1:t}) = 
\sigma^k(\vec{h}_{t,j})$,
where $\sigma : \mathbb{R}^{K+1} \rightarrow \mathbb{R}^{K+1}$ satisfies:
\begin{equation}
\label{eq:softmax}
\begin{aligned}
&\sigma(\vec{z}) = \brl{\sigma^0(\vec{z}), \dots, \sigma^K(\vec{z})}^\top, \\
&\sigma^k(\vec{z}) = \frac{\exp{(z^k)}}{\sum_{k' \in \mathcal{K}} \exp{(z^{k'})}}, \\
&\log \frac{\sigma^k(\vec{z})}{\sigma^{k'}(\vec{z})} = z^k - z^{k'} \;.
\end{aligned}
\end{equation}
To complete the Bayesian update in \eqref{eq:bayesian_update}, we propose a parametric inverse observation model, $\mathbb{P}(m^j = k | \vec{x}_t, (\vec{p}_l, \vec{y}_l))$, relating the class likelihood of map cell $m^j$ to a labeled point $\prl{\vec{p}_l, \vec{y}_l}$ obtained from state $\vec{x}_t$. 

%If the sensor ray from $\vec{x}$ toward $\vec{p}$ does not intersect $m^j$, then the observation $\prl{\vec{p}_l, \vec{y}_l}$ provides no information about the cell. In this case, we define $\mathbb{P}(m^j = k | \vec{x}, (\vec{p}_l, \vec{y}_l)) = \sigma^k(\vec{h}_{0,j})$ in terms of the prior log-odds vector $\vec{h}_{0,j}$ of cell $m^j$ (e.g., $\vec{h}_{0,j} = \vec{0}$ specifies a uniform prior over the semantic classes). Otherwise, if $m^j$ is potentially observed by the sensor ray, we define the inverse observation model as follows.

\begin{definition}
\label{def:inverse_observation_model}
Consider a labeled point $\prl{\vec{p}_l, \vec{y}_l}$ observed from state $\vec{x}_t$. Let $\mathcal{J}_{t,l} \subset \crl{1,\ldots,J}$ be the set of map cells intersected by the sensor ray from $\vec{x}_t$ toward $\vec{p}_l$. Let $m^j$ be an arbitrary map cell and $d(\vec{x}, m^j)$ be the distance between $\vec{x}$ and the center of mass of $m^j$. Define the inverse observation model of the class label of cell $m^j$ as:
\begin{equation}
\label{eq:simple_inv_obs}
\begin{aligned}
\mathbb{P}(m^j &= k | \vec{x}_t, (\vec{p}_l, \vec{y}_l)) \\
&= \begin{cases}
  \sigma^k(\tmm \bar{\vec{y}}_l \delta p_{t,l,j}), & \delta p_{t,l,j} \leq\epsilon, j \in \mathcal{J}_{t,l}\\
  \sigma^k(\vec{h}_{0,j}), & \text{otherwise},
\end{cases}
\end{aligned}
\end{equation}
where $\tmm \in \mathbb{R}^{(K+1)\times (K+1)}$ is a learnable parameter matrix, $\delta p_{t,l,j} := d(\vec{x}_t, m^j) - \norm{\vec{p}_l-\vec{x}_t}_2$, $\epsilon > 0$ is a hyperparameter (e.g., set to half the size of a cell), and $\bar{\vec{y}_l} := \brl{0, \vec{y}_l^\top}^\top$ is augmented with a trivial observation for the ``free'' class.
\end{definition}

% More intuitive explanation about the inverse observation model.
Intuitively, the inverse observation model specifies that cells intersected by the sensor ray are updated according to their distance to the ray endpoint and the detected semantic class probability, while the class likelihoods of other cells remain unchanged and equal to the prior. For example, if $m^j$ is intersected, the likelihood of the class label is determined by a softmax squashing of a linear transformation of the measurement vector $\vec{y}_l$ with parameters $\tmm$, scaled by the distance $\delta p_{t,l,j}$. Otherwise, Def.~\ref{def:inverse_observation_model} specifies an uninformative class likelihood in terms of the prior log-odds vector $\vec{h}_{0,j}$ of cell $m^j$ (e.g., $\vec{h}_{0,j} = \vec{0}$ specifies a uniform prior over the semantic classes).

\begin{definition}
The log-odds vector of the inverse observation model associated with cell $m^j$ and point observation $(\vec{p}_l, \vec{y}_l)$ from state $\vec{x}_t$ is $\vec{g}_j(\vec{x}_t, (\vec{p}_l, \vec{y}_l))$ with elements: 
\begin{align}
g^k_j(\vec{x}_t, (\vec{p}_l, \vec{y}_l)) = \log\frac{\mathbb{P}(m^j = k | \vec{x}_t, (\vec{p}_l, \vec{y}_l))}{\mathbb{P}(m^j = 0| \vec{x}_{t},(\vec{p}_l, \vec{y}_l))} \;.
\end{align}
\end{definition}

The log-odds vector of the inverse observation model, $\vec{g}_j$, specifies the increment for the Bayesian update of the cell log-odds $\vec{h}_{t,j}$ in \eqref{eq:bayesian_update}. Using the softmax properties in \eqref{eq:softmax} and Def.~\ref{def:inverse_observation_model}, we can express $\vec{g}_j$ as:
\begin{gather}
\label{eq:simple_inverse_sensor_log_odds}
\vec{g}_j(\vec{x}_t, (\vec{p}_l, \vec{y}_l)) = \begin{cases}
\tmm \bar{\vec{y}}_l \delta p_{t,l,j}, & \delta p_{t,l,j} \leq\epsilon, j \in \mathcal{J}_{t,l}\\
\vec{h}_{0,j},  & \text{otherwise}.
\end{cases}
\raisetag{3ex}
\end{gather}
%
% NN log-odds
Noting that the inverse observation model definition in \eqref{eq:simple_inv_obs} resembles a single neural network layer. One can also specify a more expressive multi-layer neural network that maps the observation $\vec{y}_l$ and the distance differential $\delta p_{t,l,j}$ along the $l$-th ray to the log-odds vector:
\begin{align}
\label{eq:nn_inverse_sensor_log_odds}
\vec{g}_j(\vec{x}_t, &(\vec{p}_l, \vec{y}_l);\tmm) \nonumber \\
&= \begin{cases}
  \textbf{NN}(\bar{\vec{y}}_l, \delta p_{t,l,j} ; \tmm) 
& \delta p_{t,l,j} \leq\epsilon, j \in \mathcal{J}_{t,l}\\
\vec{h}_{0,j}
& \text{otherwise}. 
\end{cases}
\end{align}

\begin{figure}[t]
\centering
\includegraphics[width=.95\linewidth]{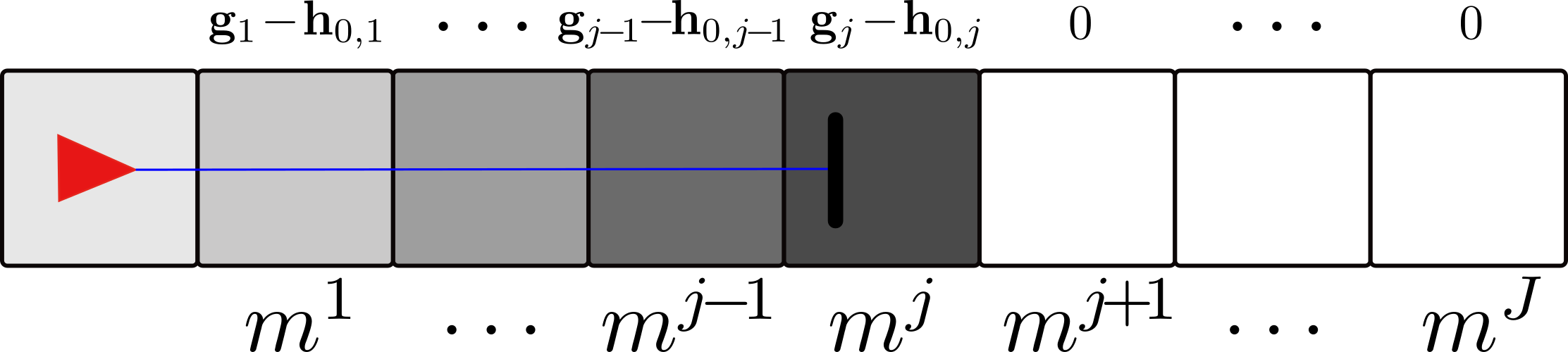}
\caption{Illustration of the log-odds update in \eqref{eq:log_odds_update} for a single point observation. The sensor ray (blue) hits an obstacle (black) in cell $m^j$. The log-odds increment $\vec{g}_j-\vec{h}_{0,j}$ on each cell is shown in grayscale.}
\label{fig:log_odds}
\end{figure}
% src/visualize_expert.py
\begin{figure}
\centering
\includegraphics[width=.95\linewidth]{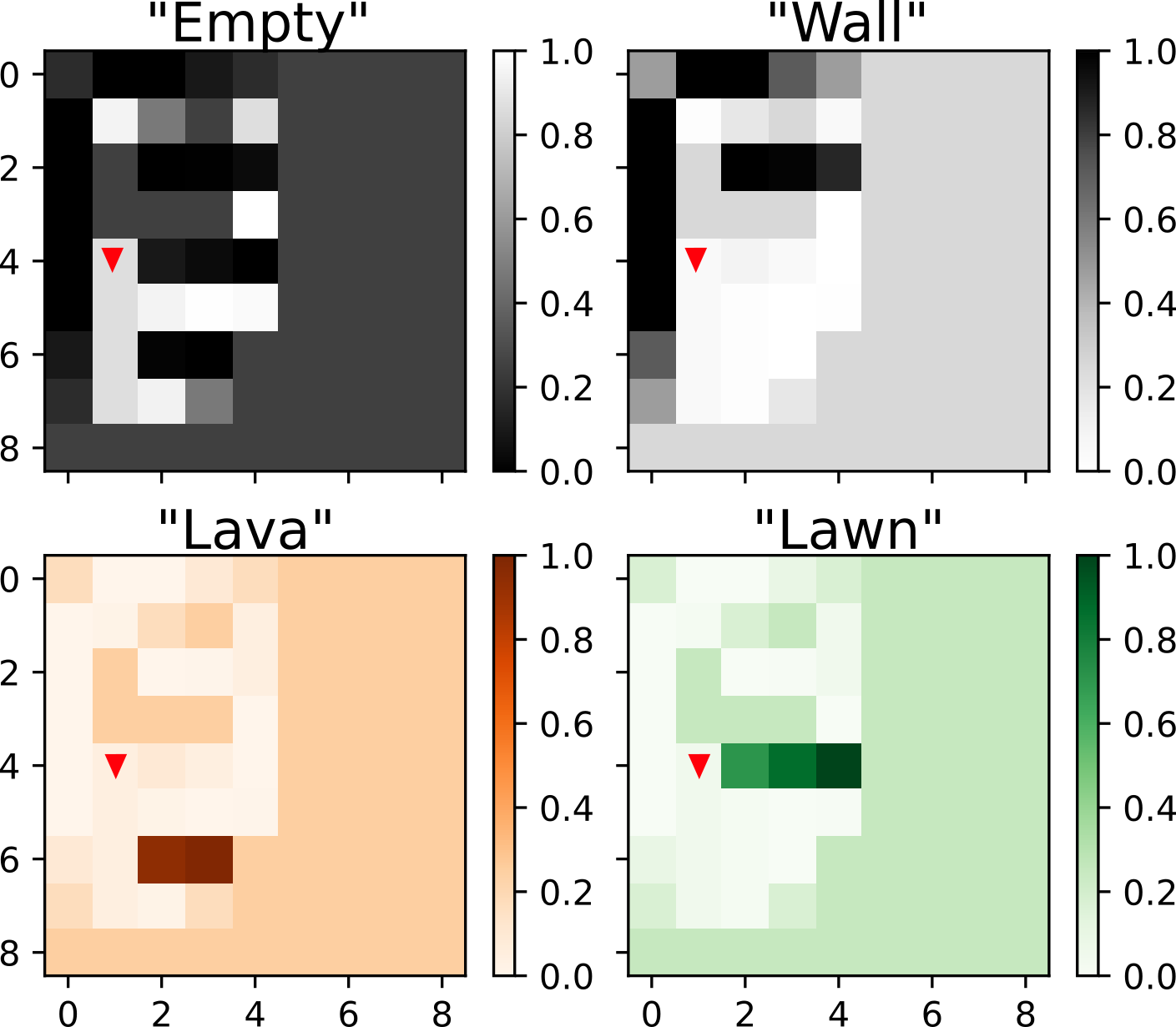}
\caption{The semantic occupancy probability of each class for the example in Fig.~\ref{fig:minigrid_env}. Using the map encoder described in Sec.~\ref{sec:map_encoder}, the semantic categories (wall, lawn, lava, etc.) can be identified correctly after training.}
\label{fig:minigrid_sem_prob}
\end{figure}

% Proposition: log-odds update
\begin{proposition}
Given a labeled point cloud $\vec{P}_{t} = \crl{\prl{\vec{p}_l, \vec{y}_l}}_l$ obtained from state $\vec{x}_t$ at time $t$, the Bayesian update of the log-odds vector of any map cell $m^j$ is:
\begin{equation}
\label{eq:log_odds_update}
\vec{h}_{t,j} = \vec{h}_{t-1,j} + \!\!\sum_{\prl{\vec{p}_l, \vec{y}_l} \in \vec{P}_{t}} \brl{\vec{g}_j(\vec{x}_t, (\vec{p}_l, \vec{y}_l)) - \vec{h}_{0,j}} \,.
\end{equation}
\end{proposition}

Fig.~\ref{fig:log_odds} illustrates the increment of the log-odds vector $\vec{h}_{t,j}$ for a single point $\prl{\vec{p}_l, \vec{y}_l}$. The log-odds of cells close to the observed point are increased more than those far away, while values of the cells beyond the observed point are unchanged. Fig.~\ref{fig:minigrid_sem_prob} shows the semantic class probability prediction for the example in Fig.~\ref{fig:minigrid_env} using the inverse observation model in Def.~\ref{def:inverse_observation_model} and the log-odds update in~\eqref{eq:log_odds_update}.

\subsection{Cost encoder}
\label{sec:cost_encoder}

\begin{figure}
\centering
\includegraphics[width=.9\linewidth]{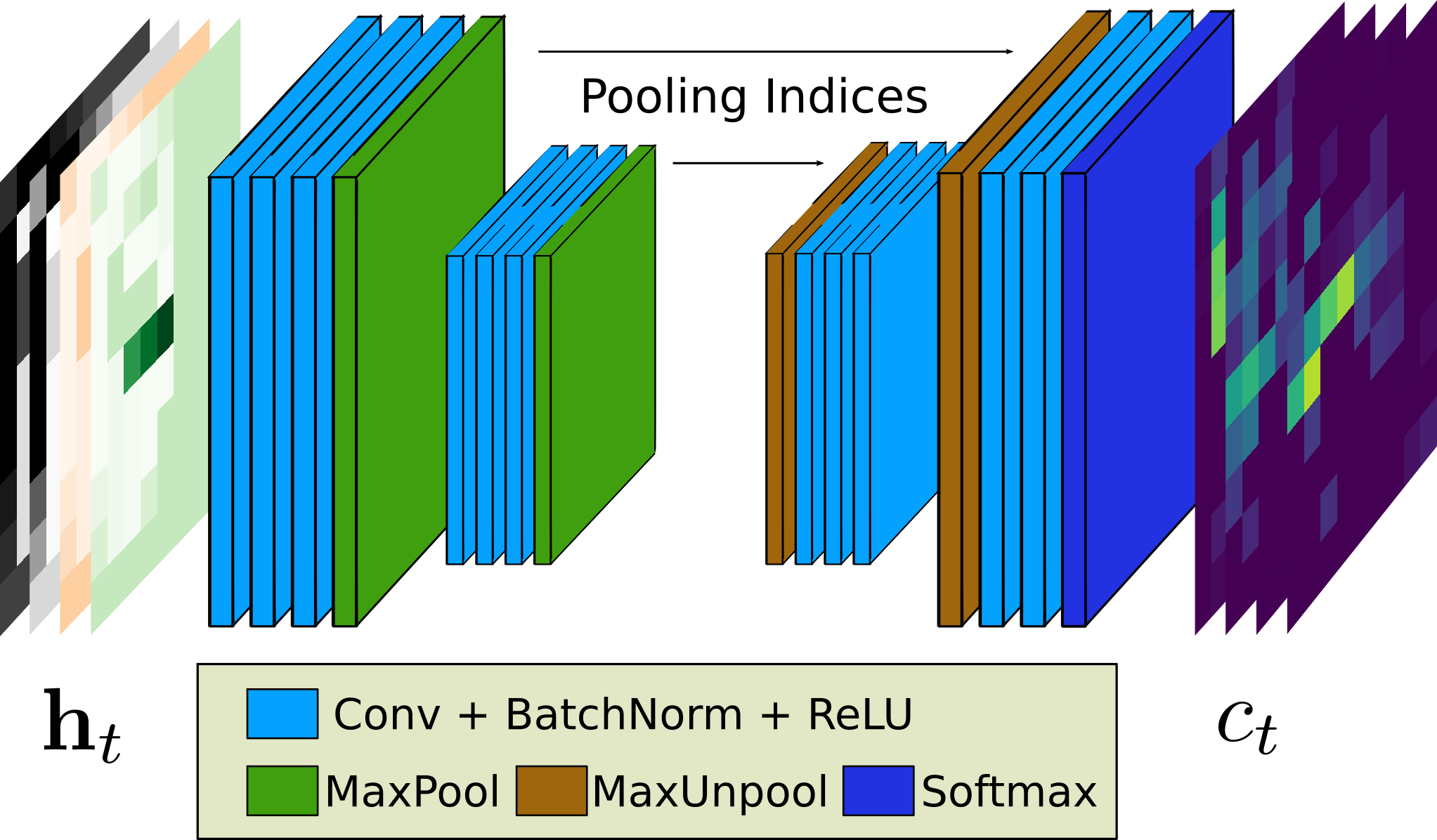}
\caption{A fully convolutional encoder-decoder neural network similar to that in \citet{Badrinarayanan2017Segnet} 
is used as the cost encoder to learn features from semantic map $\vec{h}_t$ to cost function $c_t$.}
\label{fig:encoder_decoder}
\end{figure}

We also develop a cost encoder that uses the semantic occupancy log odds $\vec{h}_{t}$ to define a cost function estimate $c_t(\vec{x},\vec{u})$ at a given state-control pair $(\vec{x},\vec{u})$. A convolutional neural network (CNN) \citep{Goodfellow-et-al-2016} with parameters $\tc$ can extract cost features from the multi-class occupancy map: $c_t = \textbf{CNN}(\vec{h}_t; \tc)$. We adopt a fully convolutional network (FCN)
architecture \citep{Badrinarayanan2017Segnet} to parameterize the cost function over the semantic class probabilities. The model is a multi-scale architecture that performs downsamples and upsamples to extract feature maps at different layers. Features from multiple scales ensure that the cost function is aware of both local and global context from the semantic map posterior.
FCNs are also translation equivariant \citep{Cohen2016GroupEquivariance}, ensuring that map regions of the same semantic class infer the same cost, irrespective of the specific locations of those regions. Our model architecture (illustrated in Fig.~\ref{fig:encoder_decoder}) consists of a series of convolutional layers with 32 channels, batch normalization \citep{Ioffe2015BatchNorm} and ReLU layers, followed by a max-pooling layer with $2\times2$ window with stride 2. The feature maps go through another series of convolutional layers with 64 channels, batch normalization, ReLU and max-pooling layers before they are upsampled by reusing the max-pooling indices. The feature maps then go through two series of upsampling, convolution, batch normalization and ReLU layers to produce the final cost function $c_t$. The final ReLU layer guarantees that the cost function is non-negative.

In summary, the semantic map encoder (parameterized by $\crl{\tmm}_l$) takes the agent state history $\vec{x}_{1:t}$ and point cloud observation history $\vec{P}_{1:t}$ as inputs to encode a semantic map probability as discussed in Sec.~\ref{sec:map_encoder}. The FCN cost encoder (parameterized by $\tc$) in turn defines a cost function from the extracted semantic features. The learnable parameters of the cost function, $c_t(\vec{x}, \vec{u}; \vec{P}_{1:t}, \vec{\theta})$, are $\vec{\theta} = \crl{\crl{\tmm}_l, \tc}$.

% src/visualize.py
\begin{figure}
\centering
\includegraphics[width=.95\linewidth]{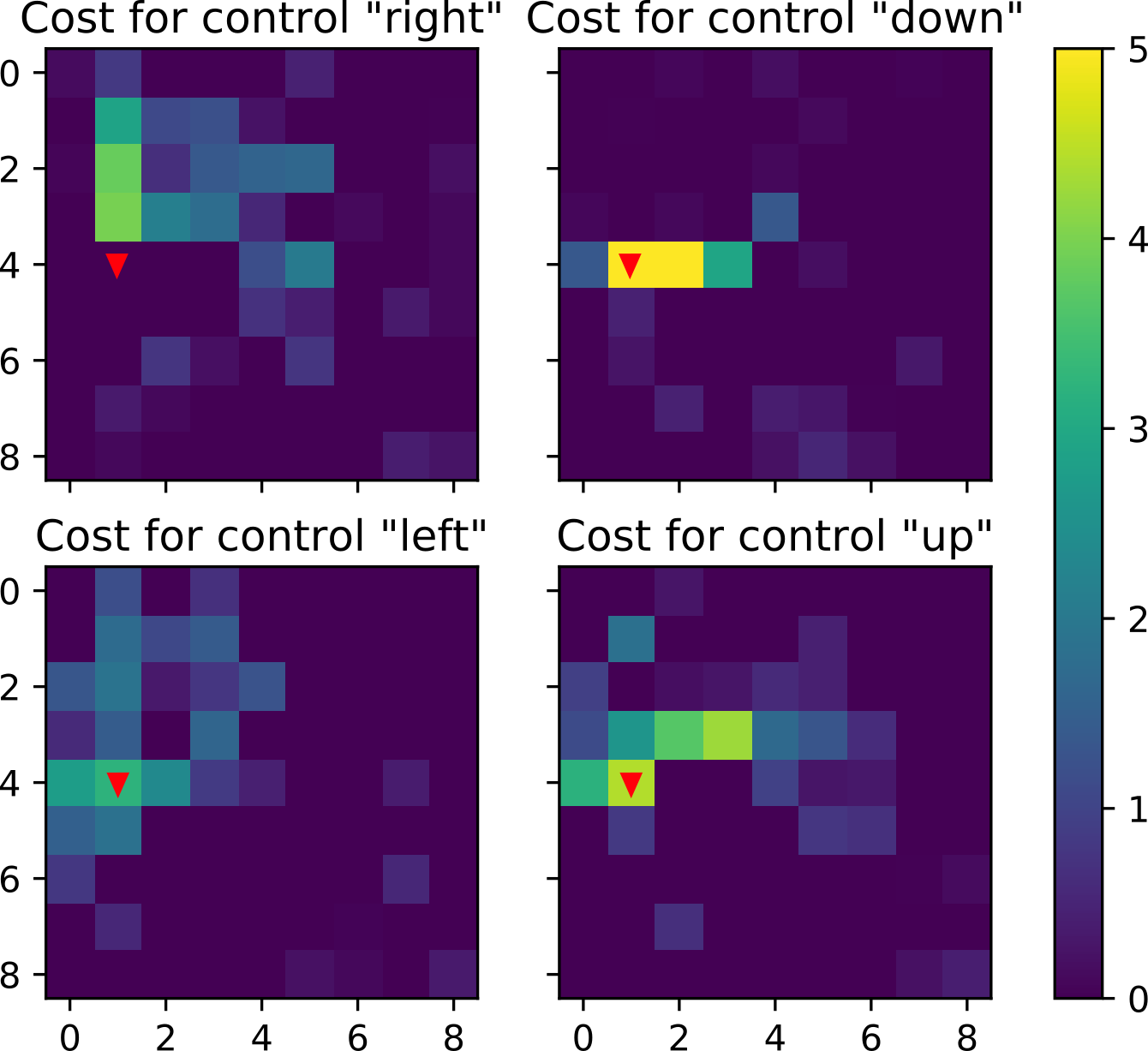}
\caption{Learned cost function for the example in Fig.~\ref{fig:minigrid_env}. The cost of control ``right" is the smallest at the agent's location after training. The agent correctly predicts that it should move right and step on the lawn.}
\label{fig:minigrid_cost}
\end{figure}

\section{Cost Learning via Differentiable Planning}
\label{sec:cost_learning}

We focus on optimizing the parameters $\vec{\theta}$ of the cost representation $c_t(\vec{x},\vec{u} ; \vec{P}_{1:t}, \vec{\theta})$ developed in Sec.~\ref{sec:cost_representation}. Since the true cost $c^*$ is not directly observable, we need to differentiate the loss function $\mathcal{L}(\vec{\theta})$ in \eqref{eq:loss}, which, in turn, requires differentiating through the DSP problem in \eqref{eq:Q} with respect to the cost function estimate $c_t$. 

Previous works rely on dynamic programming to solve the DSP problem in~\eqref{eq:Q}. For example, the VIN model \citep{Tamar2016VIN} approximates $T$ iterations of the value iteration algorithm by a neural network with $T$ convolutional and minpooling layers. This allows VIN to be differentiable with respect to the stage cost $c_t$ but it scales poorly with the size of the problem due to the full Bellman backups (convolutions and minpooling) over the state and control space. We observe that it is not necessary to determine the optimal cost-to-go $Q_t(\vec{x}, \vec{u})$ at \textit{every} state $\vec{x}\in \mathcal{X}$ and control $\vec{u} \in \mathcal{U}$. Instead of dynamic programming, a motion planning algorithm, such as a variant of A* \citep{ARA} or RRT \citep{Lavalle1998RRT, Karaman2011RRTstar}, may be used to solve problem \eqref{eq:Q} efficiently and determine the optimal cost-to-go $Q_t(\vec{x},\vec{u})$ only over a subset of promising states. The subgradient method of \citet{Shor2012Subgradient,Ratliff2006MMP} may then be employed to obtain the subgradient of $Q_t(\vec{x}_t,\vec{u}_t)$ with respect to $c_t$ along the optimal path.

\subsection{Deterministic shortest path}
\label{sec:planning}
%A backwards A* search applied to problem~\eqref{eqn:Q} with start state $\vec{x}_g$, 
%goal state $\vec{x} \in \mathcal{X}$, and predecessors expansions according to 
%transition $f$ provides an upper bound to the optimal cost-to-go: 
%\begin{align}
%Q_t^*(\vec{x},\vec{u}) &\leq c_t(\vec{x},\vec{u}) + g(f(\vec{x},\vec{u})) \nonumber\\
%&\;\;\;\forall f(\vec{x},\vec{u}) \in \text{CLOSED} \cup \text{OPEN}
%\end{align}
%where $g$ are the values computed by A* for expanded states in the CLOSED list 
%and visited states in the OPEN list. 
%Strict equality is obtained only if $f(\vec{x},\vec{u})$ belongs to the CLOSED list. 
%A Boltzmann policy $\pi_t(\vec{u} \mid \vec{x})$ may be defined using the $g$-values 
%for all $\vec{x} \in \text{CLOSED} \cup \text{OPEN} \subseteq \mathcal{X}$ and a 
%uniform distribution over $\mathcal{U}$ for all other states.

% The algorithm computes an approximation $g(\vec{x})$ of the optimal value at each state, which staisfies the following invariant at any time throughout the search:
%
%\begin{equation*}
%\begin{aligned}
%Q_t(\vec{x},\vec{u}) &= c_t(\vec{x},\vec{u}) + g(f(\vec{x},\vec{u})), \forall f(\vec{x},\vec{u}) \in CLOSED, \\ 
%Q_t(\vec{x},\vec{u}) &\leq c_t(\vec{x},\vec{u}) + g(f(\vec{x},\vec{u})), \forall f(\vec{x},\vec{u}) \notin CLOSED.
%\end{aligned}
%\end{equation*}
%

Given a cost estimate $c_t$, we use the A* algorithm (Alg.~\ref{alg:Astar}) to solve the DSP problem in \eqref{eq:Q} and obtain the optimal cost-to-go $Q_t$. The algorithm starts the search from the goal state $\vec{x}_g$ and proceeds backwards towards the current state $\vec{x}_t$. It maintains an $OPEN$ set of states, which may potentially lie along a shortest path, and a $CLOSED$ list of states, whose optimal value $\min_{\vec{u}} Q_t(\vec{x},\vec{u})$ has been determined exactly. At each iteration, the algorithm pops a state $\vec{x}$ from $OPEN$ with the smallest $g(\vec{x}) + \epsilon h(\vec{x}_t, \vec{x})$ value, where $g(\vec{x})$ is an estimate of the cost-to-go from $\vec{x}$ to $\vec{x}_g$ and $h(\vec{x}_t, \vec{x})$ is a heuristic function that does not overestimate the true cost from $\vec{x}_t$ to $\vec{x}$ and satisfies the triangle inequality. We find all predecessor states $\vec{x}'$ and their corresponding control $\vec{u}'$ that lead to $\vec{x}$ under the known dynamics model $\vec{x} = f(\vec{x}', \vec{u}')$ and update their $g$ values if there is a lower cost trajectory from $\vec{x}'$ to $\vec{x}_g$ through $\vec{x}$. The algorithm terminates when all neighbors of the current state $\vec{x}_t$ are in the $CLOSED$ set. The following relations are satisfied at any time throughout the search:
\begin{equation*}
\begin{aligned}
Q_t(\vec{x},\vec{u}) &= c_t(\vec{x},\vec{u}) + g(f(\vec{x},\vec{u})), \forall f(\vec{x},\vec{u}) \in CLOSED, \\ 
Q_t(\vec{x},\vec{u}) &\leq c_t(\vec{x},\vec{u}) + g(f(\vec{x},\vec{u})), \forall f(\vec{x},\vec{u}) \notin CLOSED.
\end{aligned}
\end{equation*}
The algorithm terminates only after all neighbors $f(\vec{x}_t, \vec{u})$ of the current state $\vec{x}_t$ are in $CLOSED$ to guarantee that the optimal cost-to-go $Q_t(\vec{x}_t,\vec{u})$ at $\vec{x}_t$ is exact. Thus, a Boltzmann policy $\pi_t(\vec{u} | \vec{x})$ can be defined using the $g$ values returned by A* for any $\vec{x} \in \mathcal{X}$:
\begin{equation}
\label{eq:planned_policy}
\pi_t(\vec{u} |\vec{x}) \propto \exp \prl{-\frac{1}{\alpha} \prl{c_t(\vec{x},\vec{u}) + g(f(\vec{x},\vec{u}))}}
\end{equation}
but only $\pi_t(\vec{u} |\vec{x}_t)$ for $\vec{u} \in \mathcal{U}$ at $\vec{x}_t$ will be needed to compute the loss function $\mathcal{L}(\vec{\theta})$ in \eqref{eq:loss} and its gradient with respect to $\vec{\theta}$ as shown next.

%and we get an upper bound to the optimal cost-to-go:
%\begin{align}
%Q_t(\vec{x},\vec{u}) &\leq c_t(\vec{x},\vec{u}; \vec{P}_{1:t}, \vec{\theta}) + g(f(\vec{x},\vec{u})) \nonumber\\
%&\;\;\;\forall f(\vec{x},\vec{u}) \in \text{CLOSED} \cup \text{OPEN} \;.
%\end{align}
%We have thus obtained a Boltzmann policy $\pi_t(\vec{u} |\vec{x}) \propto \exp (-\alpha Q_t(\vec{x}, \vec{u}))$ defined by solving the DSP~\eqref{eq:Q} with the A* algorithm on the cost estimate $c_t(\vec{x}, \vec{u}; \vec{P}_{1:t}, \vec{\theta})$. 
%Note that since all neighbors $f(\vec{x}_t, \vec{u})$ of the current state are in $CLOSED$, the optimal cost-to-go at $\vec{x}_t$ is exact:
%\begin{equation}
%Q_t(\vec{x}_t,\vec{u}) = c_t(\vec{x}_t,\vec{u}; \vec{P}_{1:t}, \vec{\theta}) + g(f(\vec{x}_t,\vec{u})) \; \forall \vec{u} \in \mathcal{U} \,.
%\end{equation}

\begin{algorithm}[t]
\DontPrintSemicolon % Some LaTeX compilers require you to use \dontprintsemicolon instead

\SetKwProg{Fn}{Function}{:}{}
\SetKwFunction{Plan}{Plan}
\Fn{\Plan{$\vec{x}_t, \vec{x}_g, c_t, h, \epsilon$}} {
$OPEN \leftarrow \crl{\vec{x}_g}, CLOSED \leftarrow \crl{}$ \;
$g(\vec{x}) \gets \infty, \forall \vec{x} \in \mathcal{X}$, $g(\vec{x}_g) \gets 0$ \;
\While{$\exists \vec{u} \in \mathcal{U} \;\text{s.t.}\; f(\vec{x}_t, \vec{u}) \notin CLOSED$  } {
  Remove $\vec{x}$ from $OPEN$ with smallest $g(\vec{x}) + \epsilon h(\vec{x}_t, \vec{x})$ and insert in $CLOSED$\;
  \For{$\prl{\vec{x}', \vec{u}'} \in$ \texttt{\upshape Predecessors}$(\vec{x})$} {
    \If{$\vec{x}' \notin CLOSED$ \textbf{\upshape and} $g(\vec{x}') > g(\vec{x}) + c_t(\vec{x}', \vec{u}') $} {
      $g(\vec{x}') \leftarrow g(\vec{x}) + c_t(\vec{x}', \vec{u}') $ \;
      $CHILD(\vec{x}') \leftarrow \vec{x}$ \;
      \eIf{$\vec{x}' \in OPEN$} {
        Update priority of $\vec{x}'$ with $g(\vec{x}') + \epsilon h(\vec{x}_t, \vec{x}')$ \;
      } {
        $OPEN \leftarrow OPEN \cup \crl{\vec{x}'}$ \;
      }
    }
  }
}
\KwRet $g(f(\vec{x}_t, \vec{u})) \;\forall \vec{u} \in \mathcal{U} $ \;
}

\SetKwFunction{Pred}{Predecessors}
\Fn{\Pred{$\vec{x}$}}{
  \KwRet $\crl{(\vec{x}', \vec{u}') \in \mathcal{X} \times \mathcal{U} \mid \vec{x} = f(\vec{x}', \vec{u}')}$\;
}

%\SetKwFunction{OptTraj}{OptimalTrajectroy}
%\Fn{\OptTraj{$\vec{x}, \vec{u}$}} {
%  $\tau = \crl{\vec{x}, \vec{u}}$\;
%  \While{$\vec{x} \neq PARENT(\vec{x})$} {
%    $\vec{x} \leftarrow PARENT(\vec{x})$ \;
%    $\tau \leftarrow \tau \cup \crl{\vec{x}, \vec{u}} \;\text{s.t.}\; f(\vec{x}, \vec{u}) = PARENT(\vec{x})$ \;
%  }
%  $\tau \leftarrow \tau \cup \crl{\vec{x}_g}$\;
%  \KwRet $\tau$\;
%}

\caption{A* motion planning}
\label{alg:Astar}
\end{algorithm}

% derived from the optimal cost-to-go in \eqref{eq:boltzmann_agent_policy} matches the demonstrations. 

\subsection{Backpropagation through planning}
\label{sec:backpropagation}

Having solved the DSP problem in \eqref{eq:Q} for a fixed cost function $c_t$, we now discuss how to optimize the cost parameters $\vec{\theta}$ such that the planned policy in \eqref{eq:planned_policy} minimizes the loss in \eqref{eq:loss}. Our goal is to compute the gradient $\frac{d \mathcal{L}(\vec{\theta})}{d \vec{\theta}}$, using the chain rule, in terms of $\frac{\partial \mathcal{L}(\vec{\theta})}{\partial Q_t(\vec{x}_t,\vec{u}_t)}$, $\frac{\partial Q_t(\vec{x}_t,\vec{u}_t)}{\partial c_t(\vec{x},\vec{u})}$, and $\frac{\partial c_t(\vec{x},\vec{u})}{\partial \vec{\theta}}$. The first gradient term can be obtained analytically from \eqref{eq:loss} and \eqref{eq:boltzmann_agent_policy}, as we show later, while the third one can be obtained via backpropagation (automatic differentiation) through the neural network cost model $c_t(\vec{x},\vec{u} ; \vec{P}_{1:t}, \vec{\theta})$ developed in Sec. \ref{sec:cost_representation}. We focus on computing the second gradient term.

We rewrite $Q_t(\vec{x}_t,\vec{u}_t)$ in a form that makes its subgradient with respect to $c_t(\vec{x},\vec{u})$ obvious. Let $\mathcal{T}(\vec{x}_t,\vec{u}_t)$ be the set of trajectories, $\vec{\tau} = \vec{x}_t, \vec{u}_t, \vec{x}_{t+1}, \vec{u}_{t+1}, \dots, \vec{x}_{T-1}, \vec{u}_{T-1}, \vec{x}_T$, of length $T$ that start at $\vec{x}_t$, $\vec{u}_t$, satisfy transitions $\vec{x}_{t+1} = f(\vec{x}_t, \vec{u}_t)$ and terminate at $\vec{x}_T = \vec{x}_g$. Let $\vec{\tau}^* \in \mathcal{T}(\vec{x}_t,\vec{u}_t)$ be an optimal trajectory corresponding to the optimal cost-to-go $Q_t(\vec{x}_t,\vec{u}_t)$. Define a state-control visitation function indicating if a transition $(\vec{x},\vec{u})$ appears in $\vec{\tau}$:
\begin{equation}
\mu_{\vec{\tau}}(\vec{x},\vec{u}) \defeq \sum_{k=t}^{T-1} \indicator_{(\vec{x}_k,\vec{u}_k) = (\vec{x},\vec{u})}.
\end{equation}
The optimal cost-to-go $Q_t(\vec{x}_t,\vec{u}_t)$ can be viewed as a minimum over  $\mathcal{T}(\vec{x}_t,\vec{u}_t)$ of the inner product of the cost function $c_t$ and the visitation function $\mu_{\vec{\tau}}$:
\begin{equation}
\label{eq:inner_product_q}
Q_t(\vec{x}_t,\vec{u}_t) = \min_{\vec{\tau} \in \mathcal{T}(\vec{x}_t,\vec{u}_t)}  
\sum_{\vec{x} \in \mathcal{X},\vec{u}\in\mathcal{U}} c_t(\vec{x},\vec{u}) 
\mu_{\vec{\tau}}(\vec{x},\vec{u}),
\end{equation}
where $\mathcal{X}$ can be assumed finite because both $T$ and $\mathcal{U}$ are finite. We make use of the subgradient method \citep{Shor2012Subgradient,Ratliff2006MMP} to compute a subgradient of $Q_t(\vec{x}_t,\vec{u}_t)$ with respect to $c_t$.

% src/visualize.py
\begin{figure}
\centering
\includegraphics[width=0.95\linewidth]{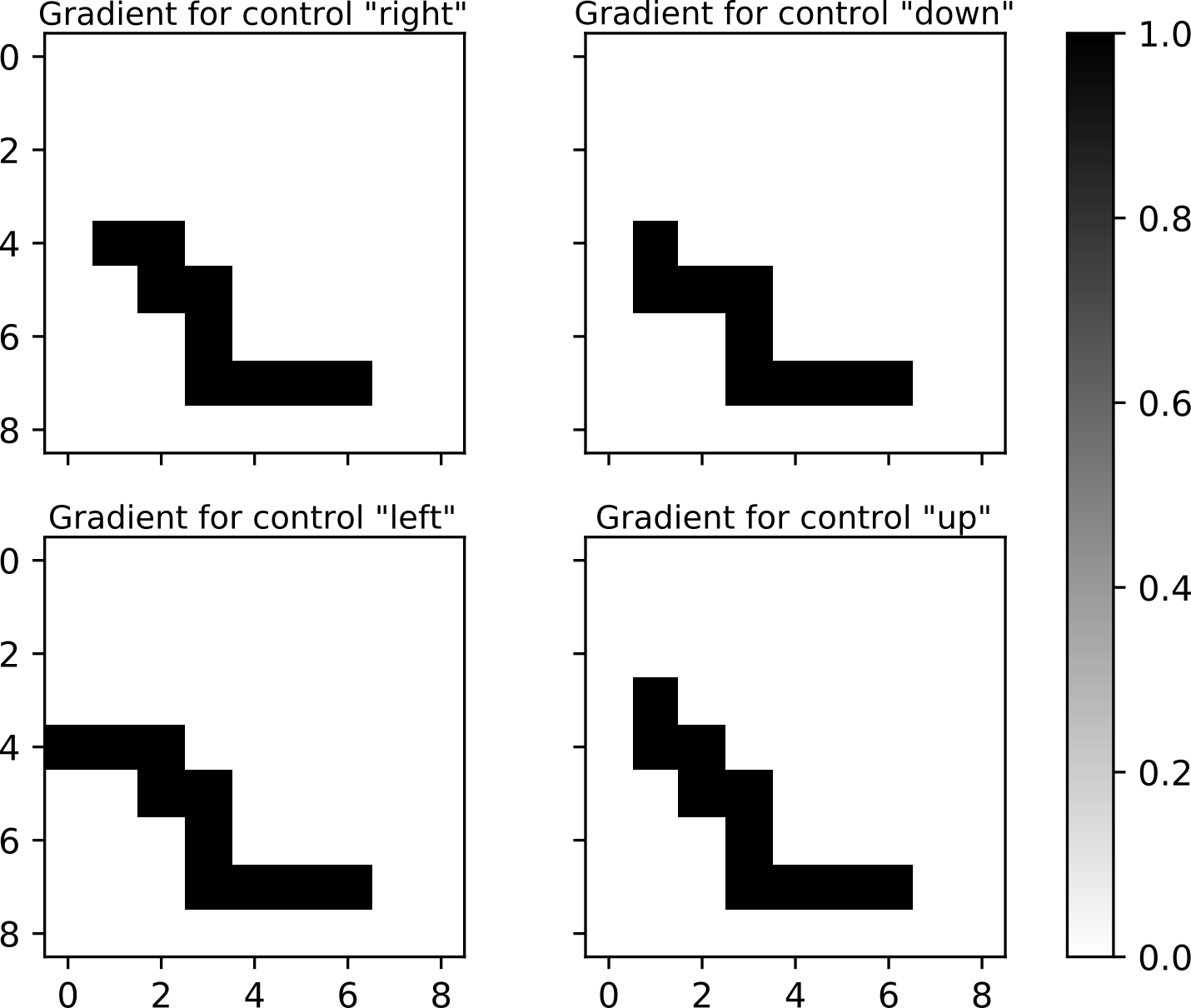}
\caption{Subgradient of the optimal cost-to-go $Q_t(\vec{x}_t,\vec{u}_t)$ for each control $\vec{u}_t$ with respect to the cost $c_t(\vec{x},\vec{u})$ in Fig.~\ref{fig:minigrid_cost}.}
\label{fig:minigrid_gradient}
\end{figure}

\begin{lemma}
\label{lemma:subgradient}
Let $f(\vec{x},\vec{y})$ be differentiable and convex in $\vec{x}$. 
Then, $\nabla_{\vec{x}} f(\vec{x}, \vec{y}^*)$, where $\vec{y}^* := \arg\min_{\vec{y}} f(\vec{x},\vec{y})$, is a subgradient of the piecewise-differentiable convex function $g(\vec{x}) := \min_{\vec{y}} f(\vec{x},\vec{y})$.
\end{lemma}

Applying Lemma~\ref{lemma:subgradient} to~\eqref{eq:inner_product_q} leads to the following subgradient of the optimal cost-to-go function:
\begin{equation}
\label{eq:subgradient_q}
\frac{\partial Q_t(\vec{x}_t,\vec{u}_t)}{\partial c_t(\vec{x},\vec{u})} 
= \mu_{\vec{\tau}^*}(\vec{x},\vec{u})
\end{equation} 
which can be obtained along the optimal trajectory $\vec{\tau}^*$ by tracing the $CHILD$ relations returned by Alg.~\ref{alg:Astar}. Fig.~\ref{fig:minigrid_gradient} shows an illustration of this subgradient computation with respect to the cost estimate in Fig.~\ref{fig:minigrid_cost} for the example in Fig.~\ref{fig:minigrid_env}. The result in \eqref{eq:subgradient_q} and the chain rule allow us to obtain a complete subgradient of $\mathcal{L}(\vec{\theta})$.

%Illustration of the subgradient from planning in Fig.~\ref{fig:minigrid_env} is shown in Fig.~\ref{fig:minigrid_gradient}.

\begin{proposition}
\label{prop:chain_rule}
A subgradient of the loss function $\mathcal{L}(\vec{\theta})$ in~\eqref{eq:loss} with respect to $\vec{\theta}$ can be obtained as:
%\begin{equation*}
%\scaleMathLine{
%\begin{aligned}
%\frac{\partial \mathcal{L}(\vec{\theta})}{\partial \vec{\theta}} 
%&= - \sum_{n=1}^N \sum_{t=1}^{T_n} \frac{\partial \log \pi_{t,n}(\vec{u}_{t,n}^* 
%\mid \vec{x}_{t,n})}{\partial \vec{\theta}}  \\
%&=- \sum_{n=1}^N \sum_{t=1}^{T_n}\sum_{\vec{u}_{t,n} \in \mathcal{U}} 
%\frac{\partial \log \pi_{t,n}(\vec{u}_{t,n}^* \mid \vec{x}_{t,n})}
%{\partial Q_{t,n}^*(\vec{x}_{t,n},\vec{u}_{t,n})} 
%\frac{\partial Q_{t,n}^*(\vec{x}_{t,n},\vec{u}_{t,n})}
%{\partial \vec{\theta}}   \\
%&=- \sum_{n=1}^N \sum_{t=1}^{T_n} 
%&\sum_{\vec{u}_{t,n} \in \mathcal{U}} \prl{\indicator_{\{\vec{u}_{t,n} = \vec{u}_{t,n}^*\}} 
%\pi_{t,n}(\vec{u}_{t,n} | \vec{x}_{t,n})} \\
%&& \times \sum_{(\vec{x},\vec{u}) \in \vec{\tau}^*} \!\!\!\!\!
%\frac{\partial Q_{t,n}^*(\vec{x}_{t,n},\vec{u}_{t,n})}{\partial c_t(\vec{x},\vec{u})} 
%\frac{\partial c_t(\vec{x},\vec{u})}{\partial \vec{\theta}}
%\end{aligned}}
%\end{equation*}
\begin{align}
\label{eq:subgradient}
&\frac{\partial \mathcal{L}(\vec{\theta})}{\partial \vec{\theta}} 
= - \sum_{n=1}^N \sum_{t=1}^{T_n} \frac{d \log \pi_{t,n}(\vec{u}_{t,n}^* 
\mid \vec{x}_{t,n})}{d \vec{\theta}}  \nonumber \\
&=- \sum_{n=1}^N \sum_{t=1}^{T_n}\sum_{\vec{u}_{t,n} \in \mathcal{U}} 
\frac{d \log \pi_{t,n}(\vec{u}_{t,n}^* \mid \vec{x}_{t,n})}
{d Q_{t,n}(\vec{x}_{t,n},\vec{u}_{t,n})} 
\frac{d Q_{t,n}(\vec{x}_{t,n},\vec{u}_{t,n})}
{d \vec{\theta}}   \nonumber \\
&=- \sum_{n=1}^N \sum_{t=1}^{T_n} 
\sum_{\vec{u}_{t,n} \in \mathcal{U}} \frac{1}{\alpha} \prl{\indicator_{\{\vec{u}_{t,n} = \vec{u}_{t,n}^*\}} -
\pi_{t,n}(\vec{u}_{t,n} | \vec{x}_{t,n})} \nonumber \\
& \;\;\;\; \times \sum_{(\vec{x},\vec{u}) \in \vec{\tau}^*} \!\!\!\!\!
\frac{\partial Q_{t,n}(\vec{x}_{t,n},\vec{u}_{t,n})}{\partial c_t(\vec{x},\vec{u})} 
\frac{\partial c_t(\vec{x},\vec{u})}{\partial \vec{\theta}}
\end{align}
\end{proposition}

\subsection{Algorithms}
\label{sec:algorithms}
The computation graph implied by Prop.~\ref{prop:chain_rule} is illustrated in Fig.~\ref{fig:approach}. The graph consists of a cost representation layer and a differentiable planning layer, allowing end-to-end minimization of $\mathcal{L}(\vec{\theta})$ via stochastic subgradient descent. The training algorithm for solving Problem~\ref{pb:1} is shown in Alg.~\ref{alg:train}. The testing algorithm that enables generalizing the learned semantic mapping and planning behavior to new sensory data in new environments is shown in Alg.~\ref{alg:test}.

\begin{algorithm}[h!]
\DontPrintSemicolon % Some LaTeX compilers require you to use \dontprintsemicolon instead
\KwIn{Dataset $\mathcal{D} \!=\! \crl{(\vec{x}_{t,n},\vec{u}_{t,n}^*,
    \vec{P}_{t,n}, \vec{x}_{g,n})}_{t=1, n=1}^{T_n,N}\!\!$}
%\KwOut{Trained cost function parameters $\theta$}
\While{$\vec{\theta}$ not converged}{
    $\mathcal{L}(\vec{\theta}) \gets 0$\;
    \For{$n = 1, \ldots,N$ \textbf{\upshape and} $t = 1,\ldots, T_n$}{
        Update $c_{t,n}$ using $\vec{x}_{t,n}$ and $\vec{P}_{t,n}$ as in 
            Sec.~\ref{sec:cost_representation}\;
        Get $Q_{t,n}(\vec{x}_{t,n},\vec{u})$ from Alg.~\ref{alg:Astar} with cost $c_{t,n}$\;
        Get $\pi_{t,n}(\vec{u} | \vec{x}_{t,n})$ in \eqref{eq:boltzmann_agent_policy} from $Q_{t,n}(\vec{x}_{t,n},\vec{u})$ \;
        $\mathcal{L}(\vec{\theta}) \gets \mathcal{L}(\vec{\theta}) -\log \pi_{t, n} 
            (\vec{u}_{t,n}^*|\vec{x}_{t,n})$\;
    }
    Update $\vec{\theta} \gets \vec{\theta} - \eta \nabla \mathcal{L}(\vec{\theta})$ via
        Prop.~\ref{prop:chain_rule}\;
}
\KwOut{Trained cost function parameters $\vec{\theta}$}
%\Return{$\vec{\theta}$} 
\caption{Train cost parameters $\vec{\theta}$}
\label{alg:train}
\end{algorithm}

\begin{algorithm}[h!]
\DontPrintSemicolon
\KwIn{Start state $\vec{x}_s$, goal state $\vec{x}_g$, cost parameters $\vec{\theta}$}
%\KwOut{Navigation succeeds or fails}
Current state $\vec{x}_t \gets \vec{x}_s$ \;
\While{$\vec{x}_t \neq \vec{x}_g$ \textbf{\upshape and} navigation \textbf{not} failed}{
    Make an observation $\vec{P}_t$\;
    Update $c_t$ using $\vec{x}_t$ and $\vec{P}_t$ as in Sec.~\ref{sec:cost_representation}\;
    Get $Q_{t}(\vec{x}_t,\vec{u})$ from Alg.~\ref{alg:Astar} with cost $c_{t}$\;
    Get $\pi_t(\vec{u} | \vec{x}_t)$ in \eqref{eq:boltzmann_agent_policy} from $Q_t(\vec{x}_t,\vec{u})$\;
    $\vec{x}_t \gets f(\vec{x}_t, \vec{u}_t)$ with $\vec{u}_t := \argmax_{\vec{u}} \pi_t(\vec{u}|\vec{x}_t)$\;
}
\KwOut{Navigation succeeds or fails at $\vec{x}_t$}
%\Return{$\vec{x}_t$}
\caption{Test control policy $\pi_t$}
\label{alg:test}
\end{algorithm}

\section{Sparse Tensor Implementation}
\label{sec:sparse_tensor}

In this section, we propose a sparse tensor implementation of the map and cost variables introduced in Sec.~\ref{sec:cost_representation}. The region explored during a single navigation trajectory is usually a small subset of the full environment due to the agent's limited sensing range. The map and cost variables $\vec{h}_t$, $\vec{g}_t$, $c_t(\vec{x},\vec{u})$ thus contains many $0$ elements corresponding to ``free'' space or unexplored regions and only a small subset of the states in $c_t(\vec{x},\vec{u})$ are queried during planning and parameter optimization in Sec.~\ref{sec:cost_learning}. Representing these variables as dense matrices is computationally and memory inefficient. Instead, we propose an implementation of the map encoder and cost encoder that exploits the sparse structure of these matrices. \citet{Choy2019MinkowskiEngine} developed the Minkowski Engine, an automatic differentiation neural network library for sparse tensors. This library is tailored for our case as we require automatic differentiation for operations among the variables $\vec{h}_t$, $\vec{g}_t$, $c_t$ in order to learn the cost parameters $\vec{\theta}$.

% sparse tensor map encoder
During training, we pre-compute the variable $\delta p_{t,l,j}$ over all points $\vec{p}_l$ from a point cloud $\vec{P}_t$ and all grid cells $m^j$. This results in a matrix $\vec{R}_t \in \mathbb{R}^{K\times J}$ where the entry corresponding to cell $m^j$ stores the vector $\vec{y}_l \delta p_{t,l,j}$\footnote{In our experiments, we found that storing only $\vec{y}_l$ at the cell $m^j$ where $p_l$ lies, instead of along the sensor ray, does not degrade performance.}. The matrix $\vec{R}_t$ is then converted to COOordinate list (COO) format \citep{Tew2016SparseTensor}, specifying the nonzero indices $\vec{C}_t \in \mathbb{R}^{N_{nz} \times 1}$ and their feature values $\vec{F}_t \in \mathbb{R}^{N_{nz} \times K}$, where $N_{nz} \ll J$ if $\vec{R}_t$ is sparse. To construct $\vec{C}_t$ and $\vec{F}_t$, we append non-zero features $\vec{y}_l\delta p_{t,l,j}$ to $\vec{F}_t$ and their coordinates $j$ in $\vec{R}_t$ to $\vec{C}_t$. The inverse observation model log-odds $\vec{g}_t$ can be computed from $\vec{C}_t$ and $\vec{F}_t$ via \eqref{eq:simple_inverse_sensor_log_odds} and represented in COO format as well. Hence, a sparse representation of the semantic occupancy log-odds $\vec{h}_t$ can be obtained 
by accumulating $\vec{g}_t$ over time via \eqref{eq:log_odds_update}.

% sparse tensor convolution
We use the sparse tensor operations (e.g., convolution, batch normalization, pooling, etc.) provided by the Minkowski Engine in place of their dense tensor counterparts in the cost encoder defined in Sec.~\ref{sec:cost_encoder}. For example, the convolution kernel does not slide sequentially over each entry in a dense tensor but is defined only over the indices in $\vec{C}_t$, skipping computations at the $0$ elements. 
% sparse tensor backpropagation 
To ensure that the sparse tensors are compatible in the backpropagtion step of the cost parameter learning (Sec.~\ref{sec:backpropagation}), the analytic subgradient in \eqref{eq:subgradient} should also be provided in sparse COO format. We implement a custom operation in which the forward function computes the cost-to-go $Q_t(\vec{x}_t, \vec{u}_t)$ from $c_t(\vec{x}, \vec{u})$ via Alg.~\ref{alg:Astar} and the backward function multiplies the sparse matrix $\frac{\partial Q_t(\vec{x}_t,\vec{u}_t)}{\partial c_t(\vec{x},\vec{u})}$ with the previous gradient in the computation graph, $\frac{\partial \mathcal{L}(\vec{\theta})}{\partial Q_t(\vec{x}_t,\vec{u}_t)}$, to get the gradient $\frac{\partial \mathcal{L}(\vec{\theta})}{\partial c_t(\vec{x},\vec{u})}$. The output gradient $\frac{\partial \mathcal{L}(\vec{\theta})}{\partial c_t(\vec{x},\vec{u})}$ is used as an input to the downstream operations defined in Sec.~\ref{sec:cost_encoder} and Sec.~\ref{sec:map_encoder} to update the cost parameters $\vec{\theta}$.

\section{MiniGrid Experiment}
\label{sec:minigrid_experiment}

We first demonstrate our inverse reinforcement learning approach in a synthetic minigrid environment \citep{gym_minigrid}. We consider a simplified setting to help visualize and understand the differentiable semantic mapping and planning components. A more realistic autonomous driving setting is demonstrated in Sec.~\ref{sec:carla_experiment}.

%We first demonstrate the efficacy of our approach in the 
%minigird environment~\citep{gym_minigrid}.
%The synthetic grid environment is a simplication of the problem setting
%to help better understand the components of our approach 
%in Sec.~\ref{sec:cost_representation} and~\ref{sec:cost_learning}.
%We provide analysis to examine the parameterization of cost function from 
%semantic occupancy probability and validate the analytic sub-gradient from planning
%the optimal path for learning the cost parameterization.

\begin{table*}[t]
  \centering
  \caption{Validation and test results for the $16\times16$ and $64\times64$ minigrid environments. We report the negative log-likelihood (\textit{NLL}) and prediction accuracy (\textit{Acc}) of the validation set expert controls and the trajectory success rate (\textit{TSR}) and modified Hausdorff distance (\textit{MHD}) between the agent and the expert trajectories on the test set. See Sec.~\ref{sec:evaluation_metrics} for precise definitions of the metrics.}
\label{tb:minigrid_test_results}
\begin{tabular}{c c c c c c c c c}
    \toprule
    &  \multicolumn{4}{c}{$16\times 16$ } & \multicolumn{4}{c}{$64\times 64$ } \\
    \cmidrule(r){2-5} \cmidrule(r){6-9}
    Model 
    & \makecell{\textit{NLL}} & \makecell{\textit{Acc} (\%)} & 
    \makecell{\textit{TSR} (\%)} & \makecell{\textit{MHD}} 
    & \makecell{\textit{NLL}} & \makecell{\textit{Acc} (\%)} & 
    \makecell{\textit{TSR} (\%)} & \makecell{\textit{MHD}} \\
    \midrule
      \texttt{DeepMaxEnt} & 0.333 & 87.7 & 85.5 & 0.783 & 0.160  & 92.5 & 86.3 & 2.305 \\
      \texttt{Ours} & 0.247 & 91.9 & 93.0 & 0.208 & 0.153 & 95.2 & 95.6 & 1.097 \\
      \bottomrule
\end{tabular}
\end{table*}

\subsection{Experiment setup}

\textit{Environment}:
Grid environments of sizes $16\times16$ and $64\times64$ are generated by sampling a random number of random length rectangles with semantic labels from $\mathcal{K} := \crl{\textit{empty}, \textit{wall}, \textit{lava}, \textit{lawn}}$. One such environment is shown in Fig.~\ref{fig:minigrid_visualization}. The agent motion is modeled over a 4-connected grid such that a control $\vec{u}_t$ from $\mathcal{U} := \crl{\textit{up}, \textit{down}, \textit{left}, \textit{right}}$ causes a transition from $\vec{x}_t$ to one of the four neighboring tiles $\vec{x}_{t+1}$. A \textit{wall} tile is not traversable and a transition to it does not change the agent's position.

%Obstacle configurations are generated randomly on maps of sizes $16\times16$ 
%and $64\times64$.
%The semantic classes of the objects in the environment includes: 
%\textit{Empty}, \textit{Wall}, \textit{Lava}, and \textit{Lawn}.
%We use a 4-connected grid such that a control $\vec{u}_t$ 
%(\textit{up}, \textit{down}, \textit{left}, \textit{right}) causes a transition 
%from $\vec{x}_t$ to one of the four neighbor tiles $\vec{x}_{t+1}$. 
%However, a \textit{Wall} tile is not traversable and a transition to a \textit{Wall} tile
%will make the robot stay at its original position.

\textit{Sensor}: 
At each step $t$, the agent receives $72$ labeled points $\vec{P}_t = \crl{\vec{p}_l,\vec{y}_l}_l$, obtained from ray-tracing a $360^{\circ}$ field of view at angular resolution of $5^{\circ}$ with maximum range of $3$ grid cells and returning the grid location $\vec{p}_l$ of the hit point and its semantic class encoded in a one-hot vector $\vec{y}_l$. See Fig.~\ref{fig:minigrid_env} for an illustration. The sensing range is smaller than the environment size, making the environment only partially observable at any given time.

\textit{Demonstrations}:
Expert demonstrations are obtained by running a shortest path algorithm on the true map $\vec{m}^*$, where the cost of arriving at an \textit{empty}, \textit{wall}, \textit{lava}, or \textit{lawn} tile is $1$, $100$, $10$, $0.5$, respectively. We generate $10000$, $1000$, and $1000$ random map configurations for training, validation, and testing, respectively. Start and goal locations are randomly assigned and maps without a feasible path are discarded.

\subsection{Models}
\texttt{DeepMaxEnt}:
We use the \texttt{DeepMaxEnt} IRL algorithm of \citet{Wulfmeier2016DeepMaxEnt} as a baseline. \texttt{DeepMaxEnt} is an extension of the MaxEnt IRL algorithm \citep{Ziebart2008MaxEnt}, which uses a deep neural network to learn a cost function directly from
LiDAR observations. In contrast to our model, \texttt{DeepMaxEnt} does not have an explicit map representation. The cost representation is a multi-scale FCN \citep{Wulfmeier2016DeepMaxEnt} adapted to the $16\times16$ and $64\times64$ domains. Value iteration over the cost matrix is approximated by a finite number of Bellman backup iterations, equal to the number of map cells. The original experiments in \citet{Wulfmeier2016DeepMaxEnt} use the mean and variance of the height of 3D LiDAR points in each cell, as well as a binary indicator of cell visibility, as input features to the FCN neural network. Since our synthetic experiments are in 2D, the point count in each grid cell is used instead of the height mean and variance. This is a fair adaptation since \citet{Wulfmeier2016DeepMaxEnt} argued that obstacles generally represent areas of larger height variance which corresponds to more points within obstacles cells for our observations. We compare against the original \texttt{DeepMaxEnt} model in Sec.~\ref{sec:carla_experiment}.

%We extend the author's implementation \citep{Wulfmeier2016DeepMaxEnt} which is a deep
%learning version of the MaxEnt IRL algorithm \citep{Ziebart2008MaxEnt}.
%It uses a neural network to learn a cost function directly from
%lidar observations without explicitly having a map representation as in our model.
%We implement the ``Multi-Scale FCN'' in \citet{Wulfmeier2016DeepMaxEnt} for the
%$16\times16$ and $64\times64$ domains. 
%Value iteration is approximated by a finite number of Bellman backup iterations,
%equal to the map size.
%The original experiments in \citet{Wulfmeier2016DeepMaxEnt} use
%the mean and variance of the height of the 3D lidar points in each cell, as well as a
%binary indicator of cell visibility, as input features to the neural network.
%Since our synthetic experiments are in 2D, the count of lidar beams
%in each grid cell is used instead of the height mean and variance.
%This is a fair adaptation since \citet{Wulfmeier2016DeepMaxEnt}
%argued that obstacles generally represent areas of larger height variance
%which means more beam counts in our observations.

\texttt{Ours}:
Our model takes as inputs the semantic point cloud $\vec{P}_t$ and the agent position $\vec{x}_t$ at each time step and updates the semantic map probability via Sec.~\ref{sec:map_encoder}. The cost encoder goes through two scales of convolution and down(up)-sampling as introduced in Sec.~\ref{sec:cost_encoder}. The models are trained using the Adam optimizer \citep{Kingma2014ADAM} in Pytorch \citep{Paszke2019Pytorch}. The neural network model training and online inference during testing are performed on an Intel i7-7700K CPU and an NVIDIA GeForce GTX 1080Ti GPU.

\subsection{Evaluation metrics}
\label{sec:evaluation_metrics}

The following metrics are used for evaluation: negative log-likelihood (\textit{NLL}) and control accuracy (\textit{Acc}) for the validation set and trajectory success rate (\textit{TSR}) and modified Hausdorff distance (\textit{MHD}) for the test set. Given learned cost parameters $\vec{\theta}^*$ and a validation set $\mathcal{D} = \crl{(\vec{x}_{t,n},\vec{u}_{t,n}^*,\vec{P}_{t,n}, \vec{x}_{g,n})}_{t=1, n=1}^{T_n, N}$, policies $\pi_{t,n}(\cdot | \vec{x}_{t,n}; \vec{P}_{1:t,n}, \vec{\theta}^*)$ are computed online via Alg.~\ref{alg:Astar} at each demonstrated state $\vec{x}_{t,n}$ and are evaluated according to:
\begin{align}
&\textit{NLL}(\vec{\theta}^*,\mathcal{D}) =  \\ 
& \;\; - \frac{1}{\sum_{n=1}^{N} T_n} \sum_{n=1, t=1}^{N, T_n} 
\log \pi_{t,n}(\vec{u}_{t,n}^* | \vec{x}_{t,n}; \vec{P}_{1:t,n}, \vec{\theta}^*)   \notag\\
&\textit{Acc}(\vec{\theta}^*,\mathcal{D}) =  \\ 
& \;\; \frac{1}{\sum_{n=1}^{N} T_n} \sum_{n=1, t=1}^{N, T_n}
\indicator_{\crl{\vec{u}_{t,n}^* = \argmax_{\vec{u}} \pi_{t,n}( \vec{u} | \vec{x}_{t,n}; 
\vec{P}_{1:t,n}, \vec{\theta}^*)}} \;.\notag
\end{align}
In the test set, the agent iteratively applies control inputs $\vec{u}_{t,n} = \argmax_{\vec{u}} \pi_{t,n}(\vec{u} | \vec{x}_{t,n}; 
\vec{P}_{1:t,n}, \vec{\theta}^*)$ as described in Alg.~\ref{alg:test}. \textit{TSR} records the success rate of the resulting trajectories, where success is defined as reaching the goal state $\vec{x}_{g,n}$ within twice the number of steps of the expert trajectory. \textit{MHD} compares how close the agent trajectories $\tau_n^A$ are from the expert trajectories $\vec{\tau}_n^E$:
\begin{align}
&\textit{MHD} (\crl{\vec{\tau}_n^A},\crl{\vec{\tau}_n^E}) = \\
& \;\;\frac{1}{N} \sum_{n=1}^N \max\biggl\{ \frac{1}{T^A}\sum_{t=1}^{T^A} d(\vec{\tau}^A_{t,n}, \vec{\tau}_n^{E}),
\frac{1}{T^E}\sum_{t=1}^{T^E} d(\vec{\tau}_{t,n}^{E}, \vec{\tau}_n^{A}) \biggr\},\notag 
\end{align}
where $d(\vec{\tau}_{t,n}^A, \vec{\tau}_n^E)$ is the minimum Euclidean distance from the state $\vec{\tau}_{t,n}^A$ at time $t$ in $\vec{\tau}_n^{A}$ to any state in $\vec{\tau}_n^{E}$.

\begin{figure*}
  \centering
  \includegraphics[width=0.94\textwidth]{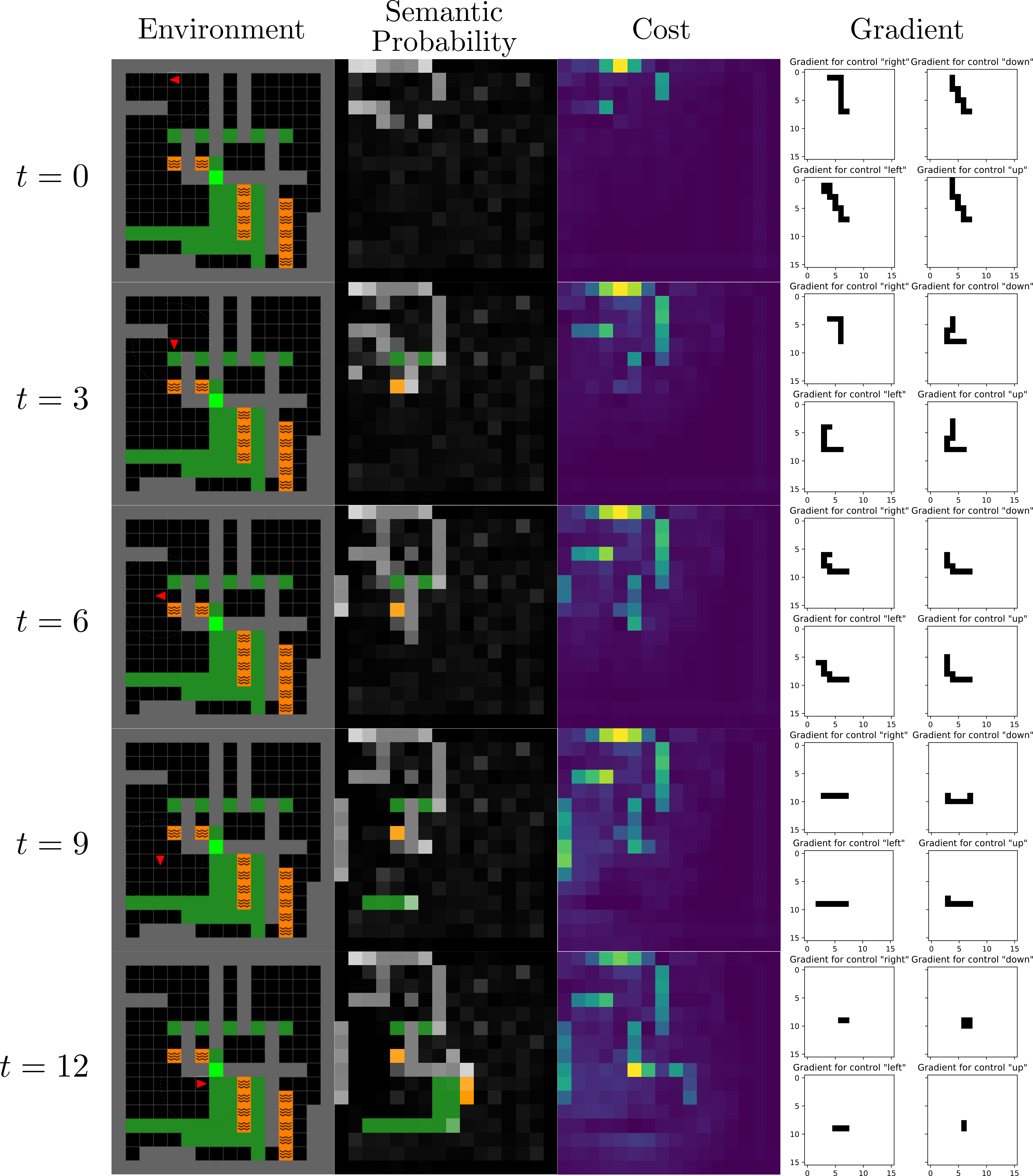}
  \caption{Examples of probabilistic multi-class occupancy estimation, cost encoder output, and subgradient computation. The first column shows the agent in the true environment at different time steps. The second column shows the semantic occupancy estimates of the different cells. The third column shows the predicted cost of arriving at each cell. Note that the learned cost function correctly assigns higher costs (in brighter scale) to \textit{wall} and \textit{lava} cells and lower costs (in darker scale) to \textit{lawn} cells. The last column shows subgradients obtained via \ref{eq:subgradient_q} during backpropagation to update the cost parameters.} 
  \label{fig:minigrid_visualization}
\end{figure*}

\subsection{Results}
The results are shown in Table.~\ref{tb:minigrid_test_results}. Our model outperforms \texttt{DeepMaxEnt} in every metric. Specifically, low \textit{NLL} on the validation set indicates that map encoder and cost encoder in our model are capable of learning a cost function that matches the expert demonstrations. During testing in unseen environment configurations, our model also achieves a higher score in successfully reaching the goal. In addition, the difference in the agent trajectory and the expert trajectory is smaller, as measured by the \textit{MHD} metric.

The outputs of our model components, i.e., map encoder, cost encoder and subgradient computation, are visualized in Fig.~\ref{fig:minigrid_visualization}. The map encoder integrates past observations and holds a correct estimate of the semantic probability of each cell. The subgradients in the last column enable us to propagate the negative log-likelihood of the expert controls back to the cost model parameters. The cost visualizations indicate that the learned cost function correctly assigns higher costs to \textit{wall} and \textit{lava} cells (in brighter scale) and lower costs to \textit{lawn} cells (in darker scale).

\subsection{Inference speed}
The problem setting in this paper requires the agent to replan at each step when a new observation $\vec{P}_t$ arrives and updates the cost function $c_t$. Our planning algorithm is computationally efficient because it searches only through a subset of promising states to obtain the optimal cost-to-go $Q_t(\vec{x}_t,\vec{u}_t)$. On the other hand, the value iteration in \texttt{DeepMaxEnt} has to perform Bellman backups on the entire state space even though most of the environment is not visited and the cost in these unexplored regions is inaccurate. Table.~\ref{tb:minigrid_inference_speed} shows the average inference speed to predict a new control $\vec{u}_t$ at each step during testing.

%Therefore, our model can generalize better to larger environment size by reducing redundant computation of the optimal cost-to-go $Q_t$ at each step.
 
%It clearly shows that \texttt{Ours} performs much faster inference in two environment scales. More importantly, \texttt{Ours} scales much better in the larger environment while inference time for \texttt{DeepMaxEnt} is more than tripled. 

\begin{table}[h]
\centering
\caption{Average inference speed comparison between our model and \texttt{DeepMaxEnt}
    for predicting one control in testing.}
\label{tb:minigrid_inference_speed}
\begin{tabular}{c c c c c}
    \toprule
    Grid size & $16\times 16$ & $64\times 64$ \\
    \midrule
    \texttt{DeepMaxEnt} & 5.8 ms  & 19.7 ms \\
    \texttt{Ours} & 2.7 ms & 3.1 ms  \\
    \bottomrule
\end{tabular}
\end{table}

\section{CARLA Experiment}
\label{sec:carla_experiment}
Building on the insights developed in the 2D minigrid environment in Sec.~\ref{sec:minigrid_experiment}, we design an experiment in a realistic autonomous driving simulation.

%We also implement a sparse tensor representation on the environment 
%based on~\citet{Choy2019MinkowskiEngine} to provide computational and memory efficiency. 

%\begin{figure}
%\centering
%\begin{subfigure}{.23\textwidth}
%  \centering
%  \includegraphics[width=\textwidth]{lidar.png}
%\end{subfigure}%
%\begin{subfigure}{.23\textwidth}
%  \centering
%  \includegraphics[width=\textwidth]{sem_cam.png}  
%\end{subfigure}
%\caption{Examples of 3D lidar points and semantic cameras facing four directions 
%in the CARLA simulator.}
%\label{fig:carla_sensors}
%\end{figure}

\subsection{Experiment setting}
\textit{Environment}:
We evaluate our approach using the CARLA simulator (0.9.9) \citep{Dosovitskiy2017CARLA}, which provides high-fidelity autonomous vehicle simulation in urban environments. Demonstration data is collected from maps $\left\{Town01, Town02, Town03\right\}$, while $Town04$ is used for validation and $Town05$ for testing. $Town05$ includes different street layouts (e.g., intersections, buildings and freeways) and is larger than the training and validation maps.

\begin{figure}
  \centering
  \includegraphics[width=\linewidth]{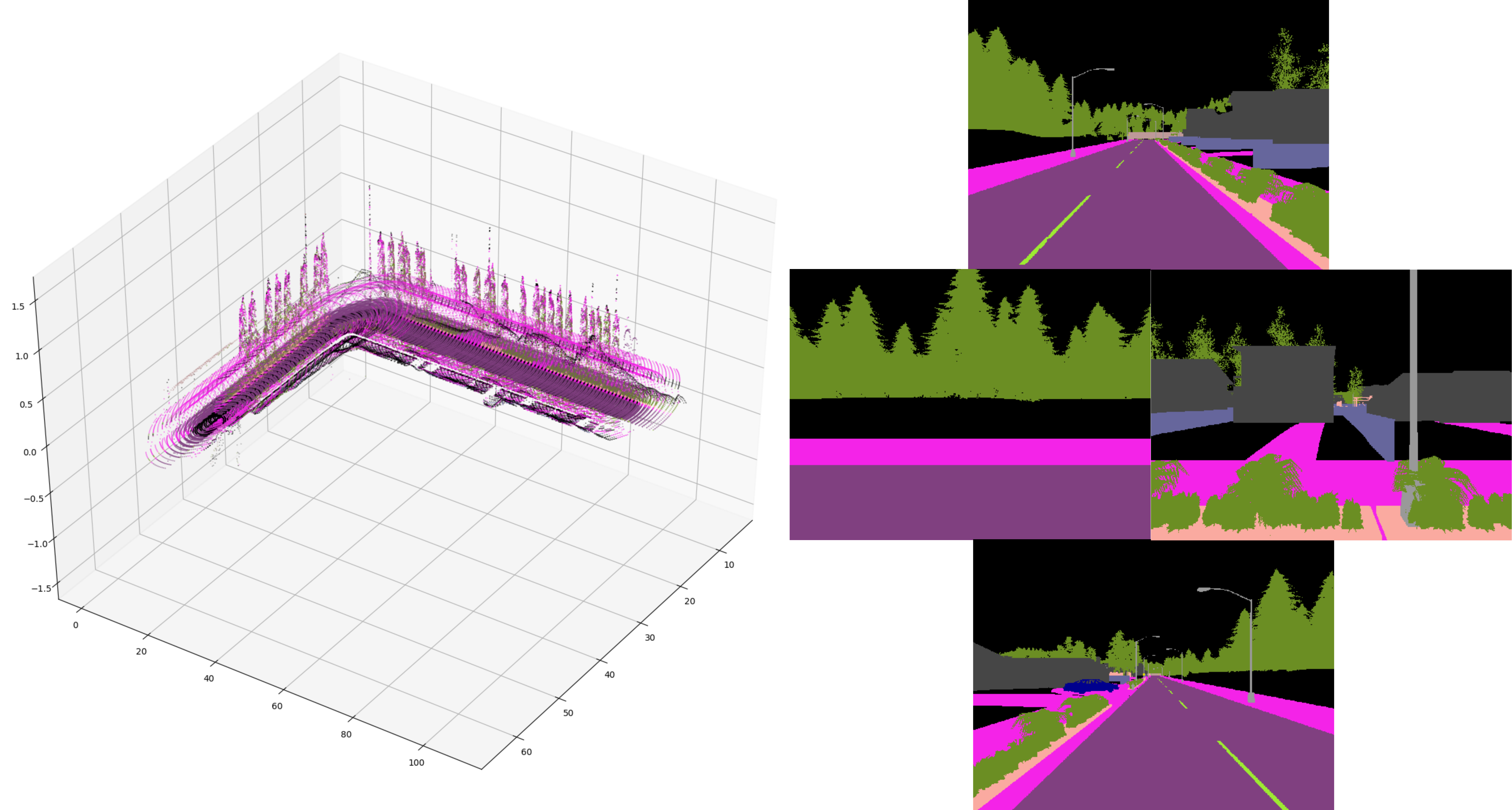}
  \caption{Example of 3D LiDAR points and semantic segmentation camera facing four directions.
  The LiDAR points are annotated with semantic class labels.}
  \label{fig:carla_sensors}
\end{figure}

\textit{Sensors}:
The vehicle is equipped with a LiDAR sensor that has $20$ meters maximum range and $360^{\circ}$ horizontal field of view. The vertical field of view ranges from $0^{\circ}$ (facing forward) to $-40^{\circ}$ (facing down) with $5^{\circ}$ resolution. A total of $56000$ LiDAR rays are generated per scan $\vec{P}_t$ and point measurements are returned only if a ray hits an obstacle (see Fig.~\ref{fig:carla_sensors}). The vehicle is also equipped with $4$ semantic segmentation cameras that detect $13$ different classes, including \textit{road}, \textit{road line}, \textit{sidewalk}, \textit{vegetation}, \textit{car}, \textit{building}, etc. The cameras face front, left, right, and rear, each capturing a $90^{\circ}$ horizontal field of view (see Fig.~\ref{fig:carla_sensors}). The semantic label of each LiDAR point is retrieved by projecting the point in the camera's frame and querying the pixel value in the segmented image.

\textit{Demonstrations}:
In each map, we collect $100$ expert trajectories by running an autonomous navigation agent provided by the CARLA Python API. On the graph of all available waypoints, the expert samples two waypoints as start and goal and searches the shortest path as a list of waypoints. The expert uses a PID controller to generate a smooth and continuous trajectory to connect the waypoints on the shortest path. The expert respects traffic rules, such as staying on the road, and keeping in the current lane. The ground plane is discretized into a $256\times256$ grid of $0.5$ meter resolution. Expert trajectories that do not fit in the given grid size are discarded. For planning purposes, the agent motion is modeled over a 4-connected grid with control space $\mathcal{U} := \crl{\textit{up}, \textit{down}, \textit{left}, \textit{right}}$. A planned sequence of such controls is followed using the CARLA PID controller. Simulation features not related to the experiment are disabled, including spawning other vehicles and pedestrians, changing traffic signals and weather conditions, etc. Designing an agent that understands more complicated environment settings with other moving objects and changing traffic lights will be considered in future research.

%To adapt to our model setting, we discretize each expert trajectory into a $128\times128$ grid of $1$ meter resolution. The agent motion is also modeled under a 4-connect grid with control space $\mathcal{U} := \crl{\textit{up}, \textit{down}, \textit{left}, \textit{right}}$. Trajectories that do not fit in the given grid size are discarded. 
%The expert respects traffic rules, such as staying on the road, and keeping in the current lane. Simulation features not related to the experiment are disabled, including spawning other vehicles and pedestrians, changing traffic signal and weather condition, etc. 
%Designing an agent that understands more complicated environment setting with other moving objects and changing traffic lights is beyond the scope of this paper and is encouraged for future research. 

\subsection{Models and metrics}
\texttt{DeepMaxEnt}:
We use the \texttt{DeepMaxEnt} IRL algorithm \citet{Wulfmeier2016DeepMaxEnt} with a multi-scale FCN cost encoder as a baseline again. Unlike the previous 2D experiment in Sec.~\ref{sec:minigrid_experiment}, we use the input format from the original paper. Specifically, observed 3D point clouds are mapped into a 2D grid with three channels: the mean and variance of the height of the points as well as the cell visibility of each cell. This model does not utilize the point cloud semantic labels. 

\texttt{DeepMaxEnt + Semantics}: 
The input features are augmented with additional channels that contain the number of points in a cell of each particular semantic class. This model uses the additional semantic information but does not explicitly map the environment over time. 

\texttt{Ours}:
We ignore the height information in the 3D point clouds $\vec{P}_{1:t}$ and maintain a 2D semantic map. The cost encoder is a two scale convolution and down(up)-sampling neural network, described in Sec.~\ref{sec:cost_encoder}. Additionally, our model is implemented using sparse tensors, described in Sec.~\ref{sec:sparse_tensor}, to take advantage of the sparsity in the map $\vec{h}_t$ and cost $c_t$. The models are implemented using the Minkowski Engine \citep{Choy2019MinkowskiEngine} and the PyTorch library \citep{Paszke2019Pytorch} and are trained with the Adam optimizer \citep{Kingma2014ADAM}. The neural network training and the online inference during testing are performed on an Intel i7-7700K CPU and an NVIDIA GeForce GTX 1080Ti GPU.

\textit{Metrics}:
The metrics, \textit{NLL}, \textit{Acc}, \textit{TSR}, and \textit{MHD}, introduced in Sec.~\ref{sec:evaluation_metrics}, are used for evaluation.

\begin{figure}
  \centering
  \includegraphics[width=\linewidth]{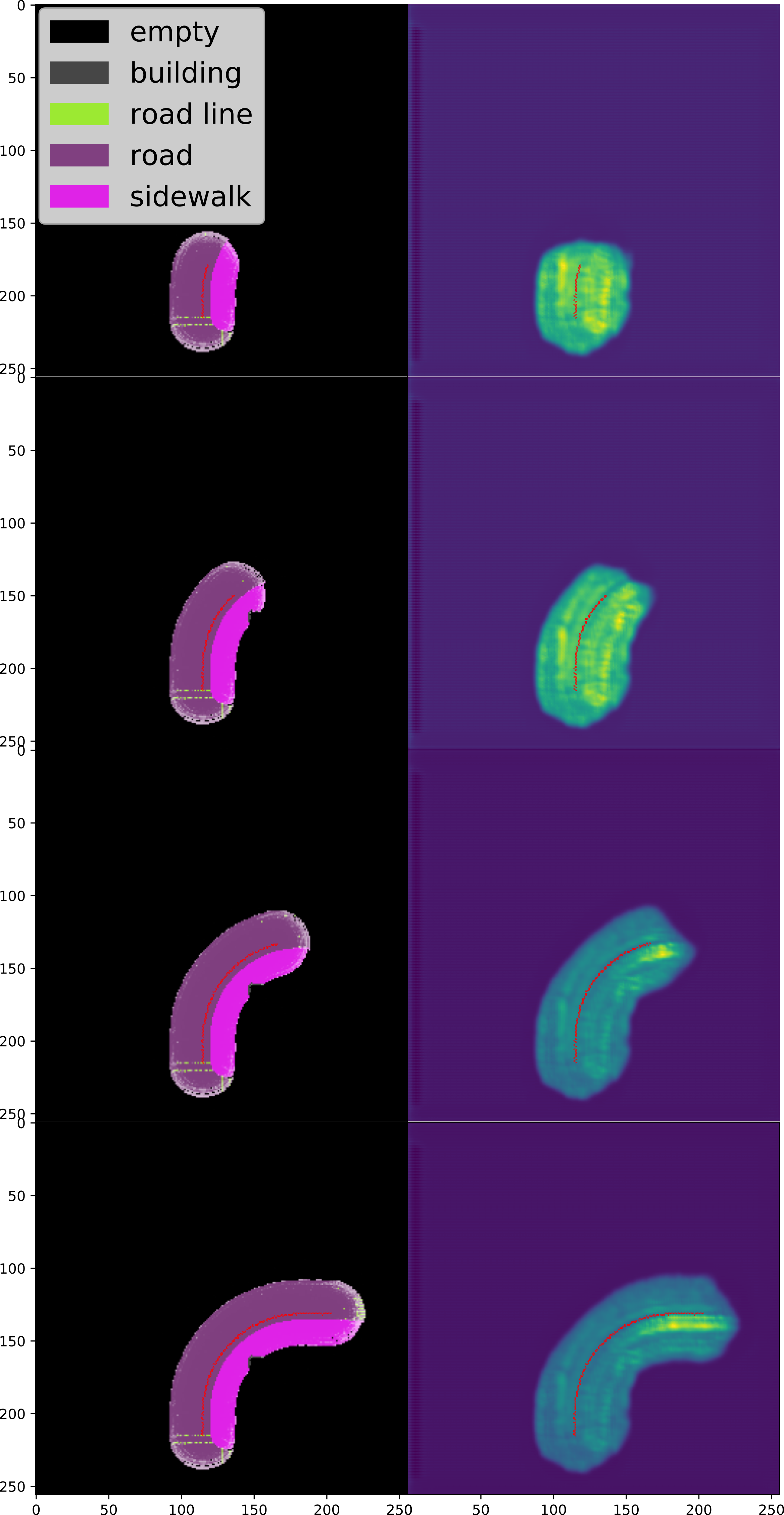}
  \caption{Examples of semantic occupancy estimation and cost encoding during different steps in a test trajectory marked in red (also see Extension 1). The left column shows the most probable semantic class of the map encoder and the right column shows the cost to arrive at each state. Our model correctly distinguishes the road from other categories (e.g., sidewalk, building, etc) and assigns lower cost to road than sidewalks.}
  \label{fig:carla_visualization}
\end{figure}

% The best model is highlighted in bold. \texttt{Ours} outperforms other methods that do not encode the semantic observation probabilistically in every metric.

\begin{table*}
  \centering
  \caption{Test results from the CARLA \textit{Town05} environment, including the negative log-likelihood (\textit{NLL}) and prediction accuracy (\textit{Acc.}) of the validation set expert controls and the trajectory success rate (\textit{TSR}) and modified Hausdorff distance (\textit{MHD}) between the agent and the expert trajectories on the test set.}
  \label{fig:carla_test_results}
  \begin{tabular}{c c c c c}
    \toprule
    Model & \makecell{\textit{NLL}} & \makecell{\textit{Acc} (\%)} & \makecell{\textit{TSR} (\%)} & \makecell{\textit{MHD}} \\
    \midrule
    \texttt{DeepMaxEnt} & 0.673 & 85.3 & 89 & 4.331 \\
    \makecell{\texttt{DeepMaxEnt + Semantics}} & 0.742 & 82.6 & 87 & 4.752 \\
    \texttt{Ours} & 0.406 & 94.2 & 93 & 2.538 \\
    \bottomrule
  \end{tabular}
\end{table*}

%\begin{figure*}[t]
%    \centering
%    \includegraphics[width=\linewidth]{fig/test_result_new.pdf}
%    \caption{
%    Example of a predicted trajectory in red at an intersection and the goal in blue.
%    The groud truth semantic map, predicted semantic map and cost map at two time steps are shown. 
%    Our model learns the sidewalk is costly to traverse.}
%    \label{fig:map_cost}
%\end{figure*}

\subsection{Results}
Table.~\ref{fig:carla_test_results} shows the performance of our model in comparison to \texttt{DeepMaxEnt} and \texttt{DeepMaxEnt + Semantics}. Our model learns to generate policies closest to the expert demonstrations in the validation map $Town04$ by scoring best in \textit{NLL} and \textit{Acc} metrics. During testing in map $Town05$, the models predict controls at each step online to generate the agent trajectory. \texttt{Ours} achieves the highest success rate of reaching the goal without hitting sidewalks and other obstacles. Among the successful trajectories, \texttt{Ours} is also closest to the expert by achieving the minimum \textit{MHD}. The results demonstrate that the map encoder captures both geometric and semantic information, allowing accurate cost estimation and generation of trajectories that match the expert behavior. Fig.~\ref{fig:carla_visualization} shows an example of a generated trajectory during testing in the previously unseen $Town05$ environment (also see Extension 1). The map encoder predicts correct semantic class labels for each cell and the cost encoder assigns higher costs to sidewalks than the road. We notice that the addition of semantic information actually degrades the performance of \texttt{DeepMaxEnt}. We conjecture that the increase in the number of input channels, due to the addition of the number of LiDAR points per category, makes the convolutional neural network layers prone to overfit on the training set but generalize poorly on the validation and test sets.

\section{Conclusion}
\label{sec:conclusion}

This paper introduced an inverse reinforcement learning approach for inferring navigation costs from demonstrations with semantic category observations. Our cost model consists of a probabilistic multi-class occupancy map and a deep fully convolutional cost encoder defined over the class likelihoods. The cost function parameters are optimized by computing the optimal cost-to-go of a deterministic shortest path problem, defining a Boltzmann control policy over the cost-to-go, and backprogating the log-likelihood of the expert controls with a closed-form subgradient. Experiments in simulated minigrid environments and the CARLA autonomous driving simulator show that our approach outperforms methods that do not encode semantic information probabilistically over time. Our work offers a promising solution for learning complex behaviors from visual observations that generalize to new environments.

%We have introduced an inverse reinforcement learning approach for infering navigation costs 
%from demonstrations with semantic observations. 
%Our model proposes a new cost representation composed of a probabilistic semantic 
%occupancy encoder and a cost encoder defined over the semantic features.
%The cost function can be optimized via backpropagation with closed-form (sub)gradient. 
%Experiments in the minigrid environment and the CARLA simulator show that our model outperforms methods that 
%do not encode semantic information probabilistically over time. 
%Our work offers a promising solution for learning semantic features in navigation 
%and may enable efficient online learning in challenging conditions.

%\begin{acks}
%We gratefully acknowledge support from NSF CRII IIS-1755568 and ONR SAI N00014-18-1-2828.
%\end{acks}

\begin{funding}
The authors disclosed receipt of the following financial support for the research, authorship, and/or publication of this article: This work was supported by the National Science Foundation [NSF CRII IIS-1755568] and the Office of Naval Research [ONR SAI N00014-18-1-2828].
\end{funding}

% References
\bibliographystyle{cls/SageH}
\bibliography{bib/ref.bib}

\begin{appendices}
\section{Index to multimedia extension}

\begin{table}[h]
\small\sf\centering
\begin{tabular}{lll}
\toprule
Extension & Media Type & Description\\
\midrule
1 & Video & \makecell{Agent rollout trajectory during\\testing in the CARLA simulator}\\
\bottomrule
\end{tabular}
\end{table}

\section{Comparison between Boltzmann and Maximum Entropy policies}
\label{sec:comparison_boltzmann_maxent_policies}

This appendix compares the MaxEnt expert model of \citet{Ziebart2008MaxEnt} to the expert model proposed in Sec.~\ref{sec:expert_model}. The MaxEnt model has been widely studied in the context of reinforcement learning and inverse reinforcement learning \citep{Haarnoja2017RLDeepEBP, Finn2016GAN_IRL_EBM, Levine2018RL_tutorial}. On the other hand, while a Boltzmann policy is a well-known method for exploration in reinforcement learning, it has not been used to model expert or learner behavior in inverse reinforcement learning.

The work of \citet{Haarnoja2017RLDeepEBP} shows that both a Boltzmann policy and the MaxEnt policy are special cases of an energy-based policy:
\begin{equation}
\label{eq:energy_based_policy}
\pi(\vec{u}_t | \vec{x}_t) \propto \exp(-E(\vec{x}_t, \vec{u}_t))
\end{equation}
with appropriate choices of the energy function $E$. We study the two policies in the discounted infinite-horizon setting, as this is the most widely used setting for the MaxEnt model. Extensions to first-exit and finite-horizon formulations are possible. Consider a Markov decision process with finite state space $\mathcal{X}$, finite control space $\mathcal{U}$, transition model $p(\vec{x}' | \vec{x}, \vec{u})$, stage cost $c(\vec{x},\vec{u})$, and discount factor $\gamma \in (0,1)$.

\begin{proposition}[{\citep[Thm.~1]{Haarnoja2017RLDeepEBP}}]
Define the maximum entropy $Q$-value as: 
\begin{align}
\label{eq:maxent_Q}
&Q_{ME} (\vec{x}_t, \vec{u}_t) := c(\vec{x}_t, \vec{u}_t)   \\
    & \;\;+ \min_{\pi} \mathbb{E}_{\pi, p} \brl{\sum_{k=t+1}^{\infty}\gamma^{k-t} \prl{c(\vec{x}_{k} , \vec{u}_{k}) 
    - \alpha \mathcal{H}(\pi(\cdot|\vec{x}_{k}))}},\notag
\end{align}
where $\mathcal{H}(\pi(\cdot|\vec{x})) = -\sum_{\vec{u} \in \mathcal{U}} \pi(\vec{u} | \vec{x}) \log \pi(\vec{u} | \vec{x})$ is the Shannon entropy of $\pi(\cdot|\vec{x})$. Then, the maximum entropy policy satisfies:
\begin{align}
\pi_{ME}(\vec{u}_t | \vec{x}_t) \propto \exp\biggl(-\frac{1}{\alpha} Q_{ME}(\vec{x}_t, \vec{u}_t)\biggr).
\end{align}
\end{proposition}

Similarly, define the usual $Q$-value as:
\begin{align}
  Q_{BM}(\vec{x}_t, \vec{u}_t) &:= c(\vec{x}_t, \vec{u}_t)  \\
  &\;\;+ \min_{\pi} \mathbb{E}_{\pi, p} \brl{\sum_{k=t+1}^{\infty}\gamma^{k-t} c(\vec{x}_{k} , \vec{u}_{k})} \notag
\end{align}
and the Boltzmann policy associated with it as:
\begin{align}
\pi_{BM}(\vec{u}_t | \vec{x}_t) \propto \exp\biggl(-\frac{1}{\alpha} Q_{BM}(\vec{x}_t, \vec{u}_t)\biggr).
\end{align}

The value functions $Q_{ME}$ and $Q_{BM}$ can be seen as the fixed points of the following Bellman contraction operators:
\begin{align}
  &\mathcal{T}_{ME}[Q](\vec{x}_t, \vec{u}_t) := c(\vec{x}_t, \vec{u}_t) \notag \\ 
    & \;\;-\gamma \alpha \mathbb{E}_{p} \brl{\log\sum_{\vec{u}_{t+1}\in \mathcal{U}}\exp\biggl(-\frac{1}{\alpha}Q(\vec{x}_{t+1}, \vec{u}_{t+1})\biggr)}\\
  &\scaleMathLine{\mathcal{T}_{BM}[Q](\vec{x}_t, \vec{u}_t) := c(\vec{x}_t, \vec{u}_t)
    + \gamma\mathbb{E}_{p} \brl{\min_{\vec{u}_{t+1} \in \mathcal{U}} Q(\vec{x}_{t+1}, \vec{u}_{t+1})}.} \notag
\end{align}
In the latter, the Q values are bootstrapped with a ``hard'' min operator, while in the former they are bootstrapped with a ``soft'' min operator given by the log-sum-exponential operation. The form of the Bellman equations resembles the online SARSA update and offline Q-learning update in reinforcement learning. Consider temporal difference control with transitions $(\vec{x}, \vec{u}, c, \vec{x}', \vec{u}')$ using SARSA backups:
\begin{equation*}
   Q(\vec{x}, \vec{u}) \leftarrow Q(\vec{x}, \vec{u}) + \eta [c(\vec{x}, \vec{u}) 
  + \gamma Q(\vec{x}', \vec{u}') - Q(\vec{x}, \vec{u})] \\
\end{equation*}
and Q-learning backups:
\begin{equation*}
   \scaleMathLine{Q(\vec{x}, \vec{u}) \leftarrow Q(\vec{x}, \vec{u}) + \eta [c(\vec{x}, \vec{u}) 
  + \gamma \min_{\vec{u}'} Q(\vec{x}', \vec{u}') - Q(\vec{x}, \vec{u})]}
\end{equation*}
where $\eta$ is a step-size parameter. If we additionally assume that the controls are sampled from the energy-based policy in \eqref{eq:energy_based_policy} defined by $Q$, the SARSA algorithm specifies the MaxEnt policy, while the Q-learning algorithm specifies the Boltzmann policy.

\begin{figure*}[ht]
  \centering
  \captionsetup[subfigure]{justification=centering}
  \begin{subfigure}{0.475\textwidth}
	\centering
	\includegraphics[width=\textwidth]{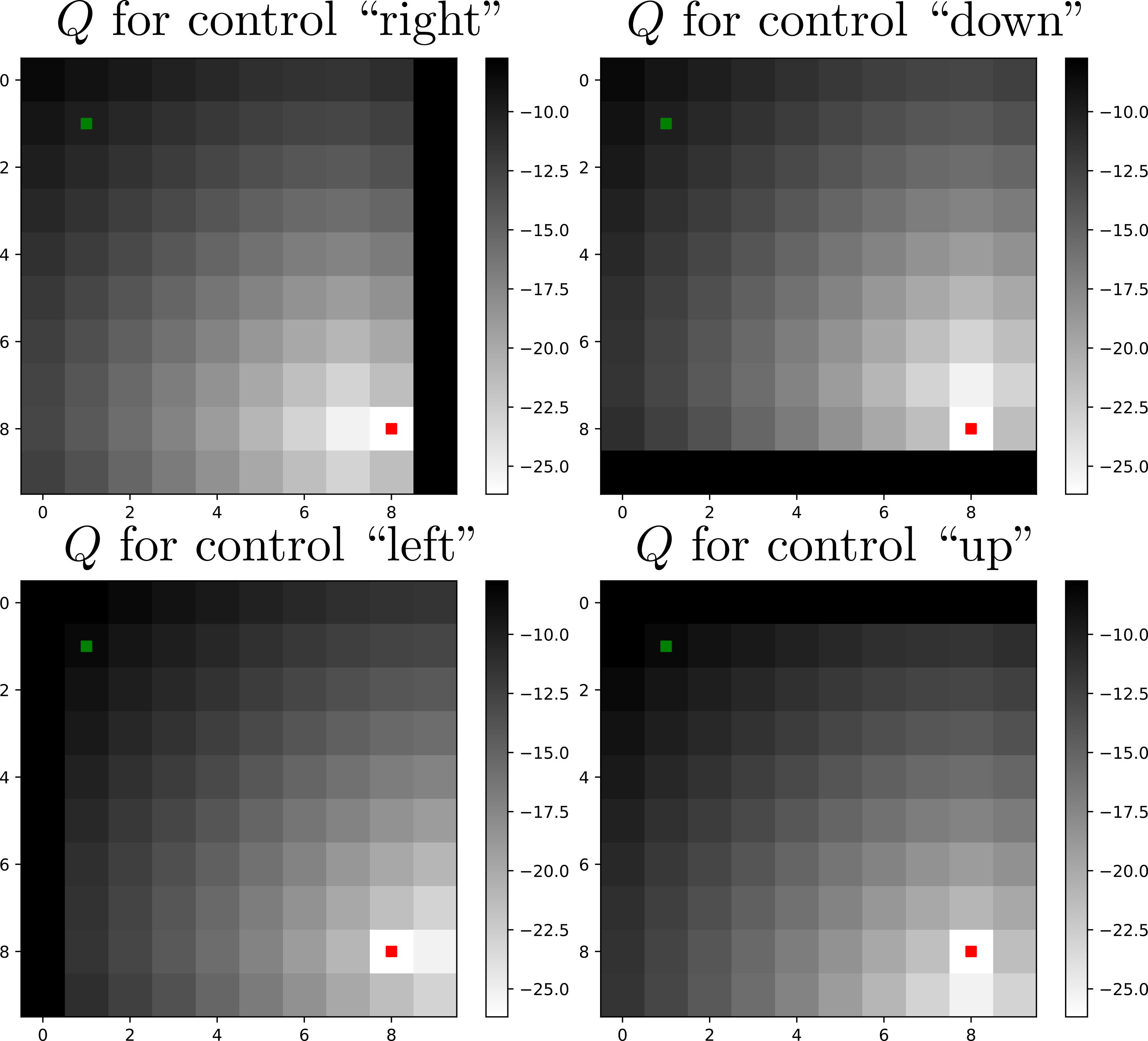}
	\caption{Value function $Q_{ME}$ for MaxEnt policy}
	\label{fig:maxent_Q}
  \end{subfigure}
  \hfill
  \begin{subfigure}{0.475\textwidth}
	\centering
	\includegraphics[width=\textwidth]{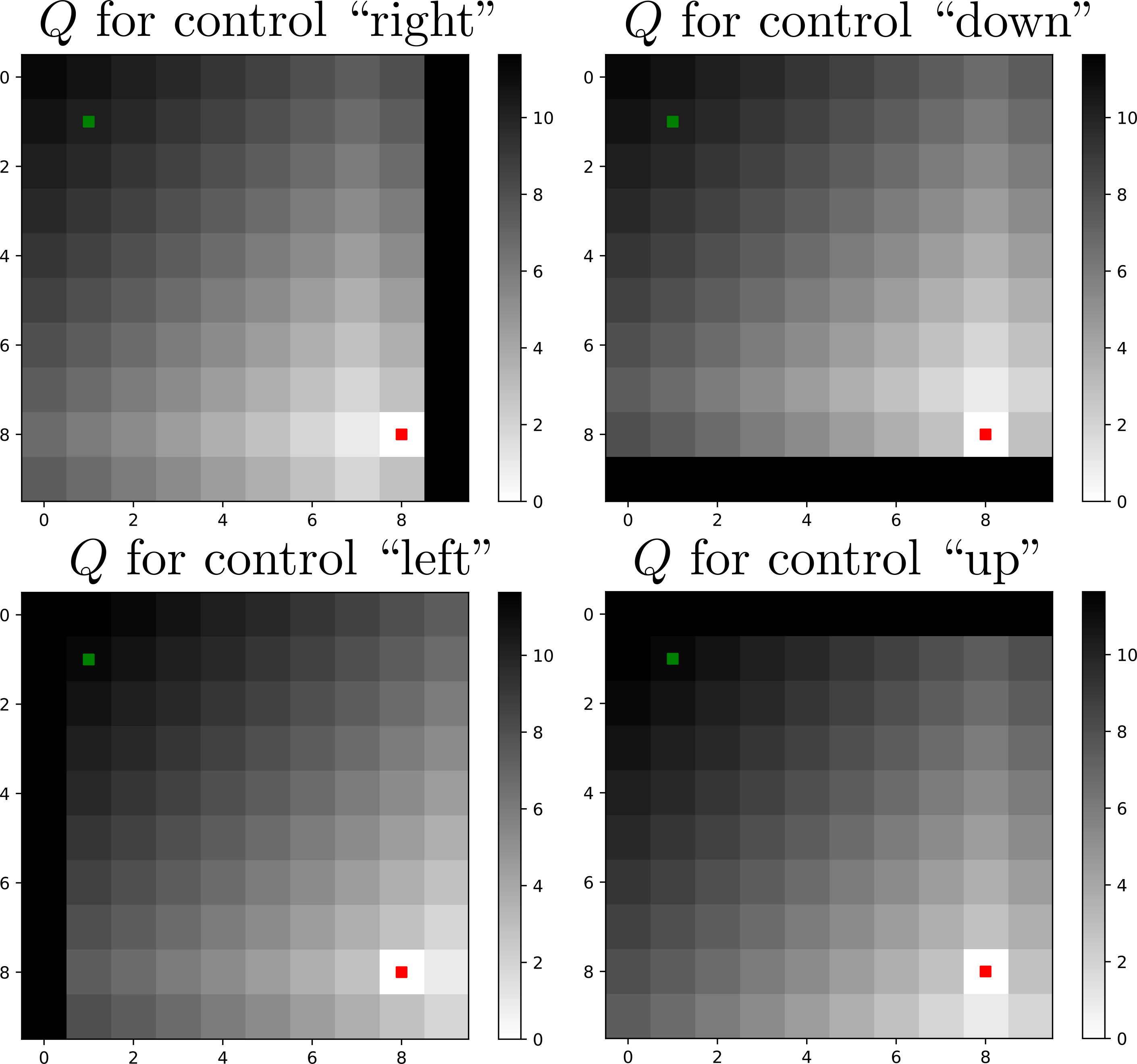}
    \caption{Value function $Q_{BM}$ for Boltzmann policy}
	\label{fig:boltzmann_Q}
  \end{subfigure}
  \vskip\baselineskip
  \begin{subfigure}{0.475\textwidth}
	\centering
	\includegraphics[width=\textwidth]{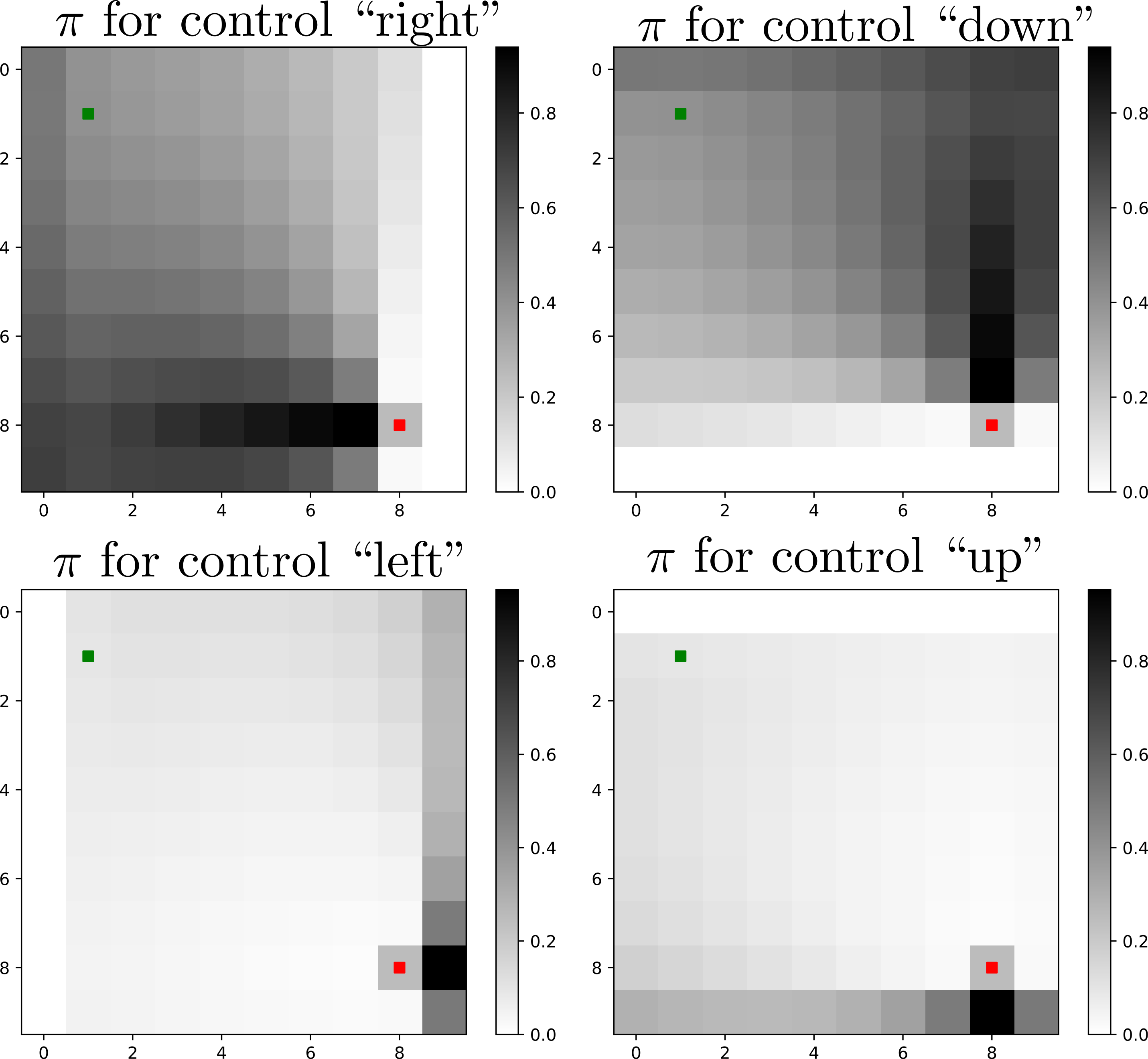}
	\caption{Maxent policy $\pi_{ME}$}
	\label{fig:maxent_pi}
  \end{subfigure}
  \hfill
  \begin{subfigure}{0.475\textwidth}
	\centering
	\includegraphics[width=\textwidth]{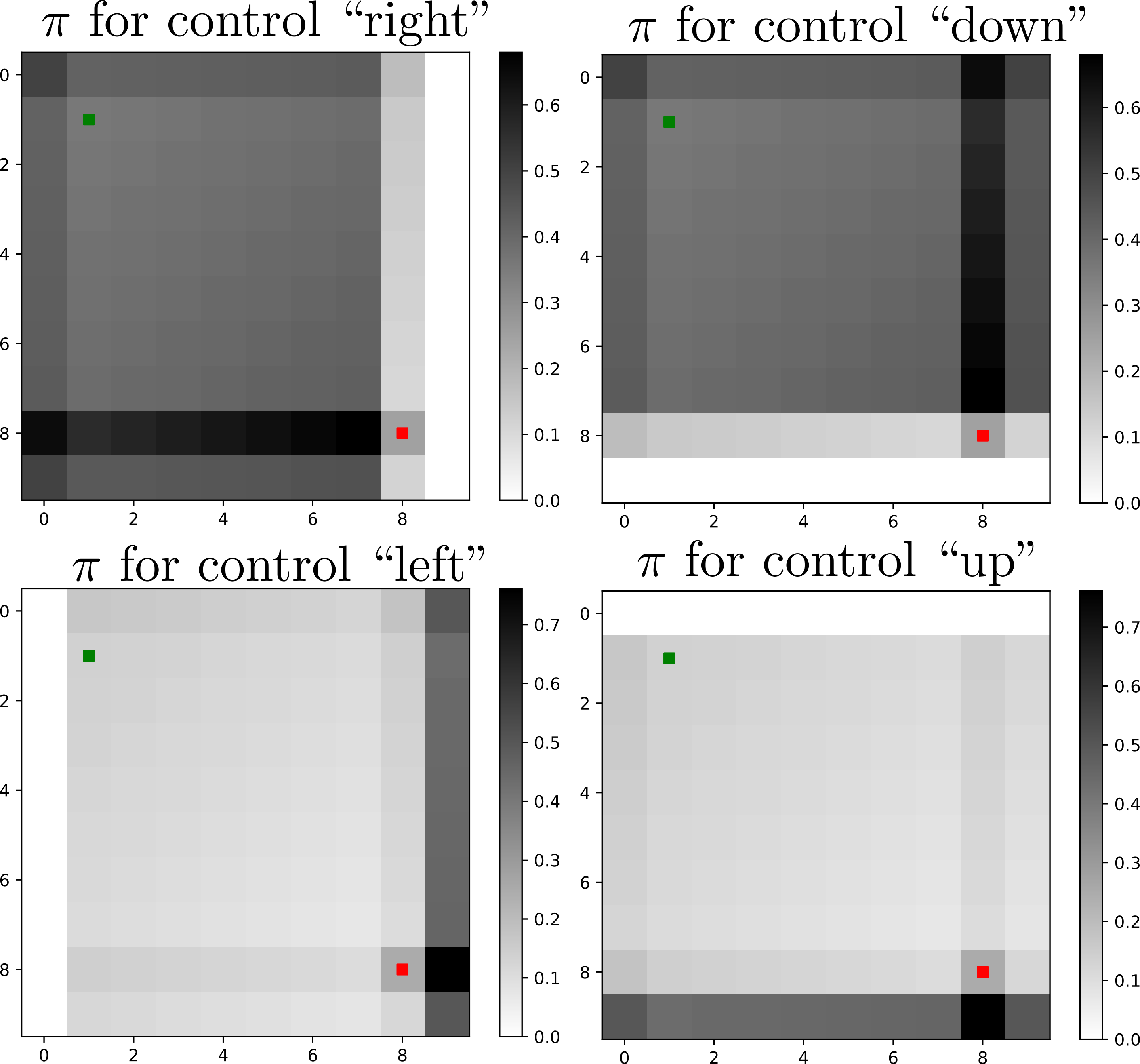}
    \caption{Boltzmann policy $\pi_{BM}$}
	\label{fig:boltzmann_pi}
  \end{subfigure}
  \caption{Value functions corresponding to the MaxEnt and Boltzmann policies in infinite horizon setting with discount $\gamma=0.95$. The environment only has obstacles around the outer boundary. The start state is marked in green and the goal in red. The controls are $\crl{\textit{right}, \textit{down}, \textit{left}, \textit{up}}$ at each state with constant true cost of $0$ to arrive at the goal (which is an absorbing state), $1$ to any state except the goal in the grid and infinity to any obstacle outside the border. Darker color indicates higher cost-to-go values to reach the goal (top two rows) or higher probability of choosing a control (bottom two rows). Although the absolute values of $Q_{ME}$ and $Q_{BM}$ are different, both have similar relative value differences across the controls, providing well-performing policies $\pi_{ME}$ and $\pi_{BM}$.}
  \label{fig:maxent_boltzmann_values_policies}
\end{figure*}

We show a visualization of the MaxEnt and Boltzmann policies, $\pi_{ME}$, $\pi_{BM}$, as well as their corresponding value functions $Q_{ME}$, $Q_{BM}$, in the infinite horizon setting with discount $\gamma=0.95$ and $\alpha=1$. The 4-connected grid environment in Fig.~\ref{fig:maxent_boltzmann_values_policies} has obstacles only along the outside border. The true cost is 0 to arrive at the goal (which is an absorbing state), 1 to any state (except the goal) inside the grid, infinity to any obstacle outside the border. Note that $Q_{ME}$ and $Q_{BM}$ are very different in absolute value. In fact, $Q_{ME}$ is negative for all states due to the additional entropy term in \eqref{eq:maxent_Q}. However, the relative value differences across the controls are similar and, thus, both policies $\pi_{ME}$ and $\pi_{BM}$ generate desirable paths from start to goal.

% Values are saturated at 0 if control incurs a next state that is out of map.

%Finally, we show a visualization of the MaxEnt and Boltzmann policies
%$\pi_{ME}, \pi_{BM}$ as well as their corresponding value functions $Q_{ME}, Q_{BM}$
%in a simple grid environment in Fig.~\ref{fig:maxent_boltzmann_values_policies}.
%The environment is a $10\times10$ grid without obstacles. 
%The start location is marked in green while the goal is in red. 
%The cost of any transition in ``right, down, left, up'' is 1, and transitions beyond the grid boundaries are not allowed.
%Note that $Q_{ME}$ and $Q_{BM}$ are very different in absolute values. 
%In fact, $Q_{ME}$ are negative for all states due to the additional entropy term 
%in Eq.~\ref{eq:maxent_Q}.
%However, the relative difference across controls are similar and thus the policies
%$\pi_{ME}$ and $\pi_{BM}$ are both desirable behaviors from start to goal.

\end{appendices}

\end{document}